%% file: egpaper.tex
\documentclass[10pt,twocolumn,letterpaper]{article}

\usepackage{iccv}
\usepackage{times}
\usepackage{epsfig}
\usepackage{graphicx}
\usepackage{amsmath}
\usepackage{amssymb}
\usepackage{color}

\usepackage[pagebackref=true,breaklinks=true,letterpaper=true,colorlinks,bookmarks=false]{hyperref}
\usepackage{caption}
\usepackage{subcaption}
\usepackage{array}
\usepackage{slashbox}
\usepackage{multirow}
\graphicspath{{./figures/}}

\iccvfinalcopy 

\def\iccvPaperID{****} 

\ificcvfinal\fi
\begin{document}

\title{A Semi-supervised Framework for Image Captioning}

\author{Wenhu Chen\\
RWTH Aachen\\
{\tt\small hustchenwenhu@gmail.com}
\and
Aurelien Lucchi\\
ETH Z\"urich\\
{\tt\small aurelien.lucchi@inf.ethz.ch}
\and
Thomas Hofmann\\
ETH Z\"urich\\
{\tt\small thomas.hofmann@inf.ethz.ch}
\and
\\
}


\newcommand{\highlight}[1]{\textcolor{red}{#1}}
\newcommand{\secondary}[1]{\textcolor{blue}{#1}}

\newcommand{\R}{\mathbb{R}}
\newcommand{\V}{\mathcal{V}}
\newcommand{\argmax}[1]{\underset{#1}{\operatorname{arg}\,\operatorname{max}}\;}
\newcommand{\width}{1.0}

\maketitle

\begin{abstract}
State-of-the-art approaches for image captioning require supervised training data consisting of captions with paired image data. These methods are typically unable to use unsupervised data such as textual data  with no corresponding images, which is a much more abundant commodity. We here propose a novel way of using such textual data by artificially generating missing visual information. We evaluate this learning approach on a newly designed model that detects visual concepts present in an image and feed them to a reviewer-decoder architecture with an attention mechanism. Unlike previous approaches that encode visual concepts using word embeddings, we instead suggest using regional image features which capture more intrinsic information. The main benefit of this architecture is that it synthesizes meaningful thought vectors that capture salient image properties and then applies a soft attentive decoder to decode the thought vectors and generate image captions. We evaluate our model on both Microsoft COCO and Flickr30K datasets and demonstrate that this model combined with our semi-supervised learning method can largely improve performance and help the model to generate more accurate and diverse captions.
\end{abstract}


\input{content/Introduction}
\input{content/Related_Works}
\input{content/Our_Model}
\input{content/Experiments}

\input{content/Conclusion}

{
\newpage
\small
\bibliographystyle{ieee}
\bibliography{egbib}
}

\onecolumn
\include{content/Appendix}

\end{document}

%% file: content/Introduction.tex

\section{Introduction}
Giving the ability to a machine to describe an image has been a long standing goal in computer vision. It proved to be an extremely challenging problem for which interest has been renewed in recent years thanks to recent developments brought by deep learning techniques. Among these are two techniques originating from computer vision and natural language processing, namely convolutional~\cite{krizhevsky2012imagenet} and recurrent neural network architectures especially Long Short-term Memory Networks~\cite{hochreiter1997long}. 

Among the most popular benchmark datasets for image captioning are MS-COCO and Flickr30K~\cite{chen2015microsoft,plummer2015flickr30k} whose recent release helped accelerating new developments in the field. In 2015, a few approaches~\cite{karpathy2015deep, mao2015learning, xu2015show,vinyals2015show,fang2015captions} set a very high standards on both datasets by reaching a BLEU-4 score of over 27\% on MS-COCO and over 19\% on Flickr30k. Most of the recent improvements have since then built on these systems and tried to figure out new ways to adapt the network architecture or improve the visual representation, e.g. using complex attention mechanisms~\cite{you2016image, yang2016encode, jin2015aligning}.

While all these developments are significant, the performance of existing state-of-the-art approaches is still largely hindered by the lack of image-caption groundtruth data. The acquisition of this training data is a painstaking process that requires long hours of manual labor~\cite{lin2014microsoft} and there is thus a strong interest in developing methods that require less groundtruth data. In this paper, we depart from the standard training paradigm and instead propose a novel training method that exploits large amounts of unsupervised text data without requiring the corresponding image content.

\begin{figure*}[ht]
\begin{center}
\fbox{\rule{0pt}{2in}
   \includegraphics[width=0.75\linewidth]{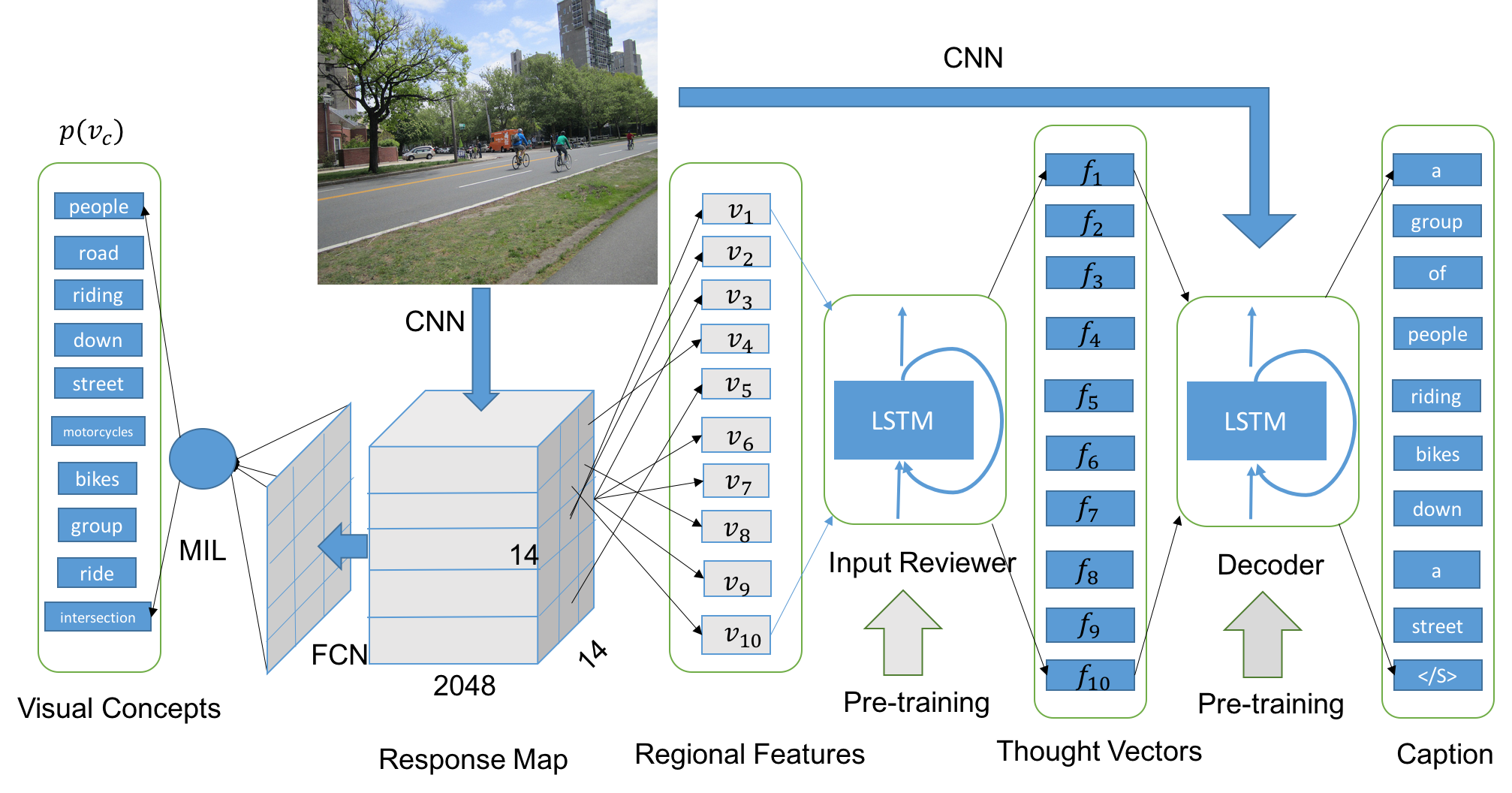}
}
   \caption{\small{Overview of our image captioning system. First, an image is fed to a CNN architecture to rank the top visual concepts appearing in this image. The feature map localizer then traces back the regions with the strongest correlation to the detected visual concepts and extract regional visual features from them. Finally, the input reviewer aggregates these regional features and produces thought vectors which are then fed to an attentive decoder to generate the correct caption.}}
   \label{fig:overview}
   \vspace{-1cm}
\end{center}
\end{figure*}

Our model is inspired by the recent success of sequence-to-sequence models in machine translation \cite{bahdanau2014neural, luong2015effective}, and multimodal recurrent structure models~\cite{xu2015show, karpathy2015deep, you2016image}. These methods encrypt images into a common vector space from which they can be decoded to a target caption space. Among these methods, ours is closely related to~\cite{you2016image} which used a fully convolutional network (FCN~\cite{ren2015faster}) and multi-instance learning (MIL~\cite{maron1998framework}) to detect visual concepts from the images. These concepts are then mapped to a word vector space and serve as input to an attention mechanism of an LSTM-based decoder for caption generation. We follow their idea but we use image region features instead of semantic word embeddings as they typically carry more salient information. Besides we add an input reviewer - as suggested in~\cite{yang2016encode} - in order to perform a given number of review steps, then output thought vectors. We feed these thought vectors to the attention mechanism of the attentive decoder. The resulting model is depicted in~\autoref{fig:overview} and relies on four building blocks: (i) a convolutional layer, (ii) a response map localizer, (iii) an attentive LSTM reviewer and (iv) an attentive decoder. All these steps will be detailed in Section~\ref{sec:model}.

Besides a novel architecture based on the work of~\cite{xu2015show, yang2016encode,you2016image}, our main contribution lies in the use of out-of-domain textual data \-- without visual data \-- to pre-train our model. We propose a novel way to artificially generate the missing visual information in order to pretrain our model. Once the model is pre-trained, we start to feed pairwise in-domain visual textual training data to fine tune the model parameters. This \textit{semi-supervised} approach yields significant improvements in terms of CIDEr~\cite{vedantam2015cider}, BLEU-4~\cite{papineni2002bleu} and METEOR~\cite{banerjee2005meteor}. Since unpaired textual data is largely available and easy to retrieve in various forms, our approach provides a novel paradigm to boost the performance of existing image captioning systems. Besides, we also made our code and pre-training dataset available on github~\footnote{\url{https://github.com/wenhuchen/ETHZ-Bootstrapped-Captioning}} to support further progress based on our approach.

%% file: content/Related_Works.tex

\section{Related Work}

\paragraph{Visual Concept Detector.}
The problem of visual concept detection has been studied in the vision community for decades. Many challenges~\cite{ILSVRC15,Everingham10,lin2014microsoft} are related to detecting objects from a given set of detectable concepts which are mainly restricted to visible entities such as "cars" or "pedestrians". Other concepts such as "sitting", "looking" or colors are typically ignored. These concepts are however important when describing the content of an image and ignoring them can thus severely hurt the performance of an image captioning system. This problem was partially addressed by the work of~\cite{fang2015captions} who proposed a weakly supervised MIL algorithm, which is able to detect broader and more abstract concepts out of the images. A similar approach was proposed in~\cite{zhou2015conceptlearner} to learn weakly labeled concepts out of a set of images.

\paragraph{Image Description Generation.}
Traditional methods for image captioning can be divided into two categories: (1) template-based methods such as~\cite{kulkarni2013babytalk} and~\cite{li2011composing}, and (2) retrieval-based methods such as~\cite{kuznetsova2012collective} and ~\cite{kuznetsova2014treetalk}. Template-based systems lack flexibility since the structure of the caption is fixed, the main task being to fill in the blanks of the predefined sentence. On the other hand, retrieval-based models heavily rely on the training data as new sentences can only be composed out of sentences coming from the training set. A recent breakthrough in image captioning came from the renewal of deep neural networks. Since then, a common theme has become utilizing both convolutional neural networks and recurrent neural networks for generating image descriptions. One of the early examples of this new paradigm is the work of~\cite{karpathy2015deep} that utilizes a deep CNN to construct an image representation, which is then fed to a bidirectional Recurrent Neural Networks. This architecture has since then become a \textit{de facto} standard in image captioning~\cite{mao2014explain, karpathy2015deep, mansimov2015generating}.

Another recent advance in the field of image captioning has been the use of attention models, initially proposed in~\cite{xu2015show} and quickly adopted by~\cite{jin2015aligning,yang2016encode} and others. These methods typically use spatially localized features computed from low layers in a CNN in order to represent fine-grained image context information while also relying on an attention mechanism that allows for salient features to dynamically become more dominant when needed. Another related approach is~\cite{you2016image} that extracts visual concepts (as in \cite{fang2015captions}) and uses an attentive LSTM to generate captions based on the embeddings of the detected visual concepts. Two attention models are then used to first synthesize the visual concepts and then to generate captions.

Among all these approaches, \cite{you2016image, xu2015show, yang2016encode} are the closest to ours in spirit. Our model borrows features from these existing systems. We for example make use of an input review module as suggested in~\cite{yang2016encode} to encode semantic embedding into richer representation of factors. We then use a soft-attention mechanism~\cite{xu2015show} to generate attention weights for each factor, and we finally use beam search to generate a caption out of the decoder.

\paragraph{Leveraging External Training Data.}
Most image captioning approaches are trained using paired image-caption data. Given that producing such data is an expensive process, there has been some interest in the community to train models with fewer data. The approach developed in~\cite{mao2015learning} allows the model to enlarge its word dictionary to describe the novel concepts using a few examples and without extensive retraining. However this approach still relies on paired data. The approach closest to ours is~\cite{hendricks2015deep} that focuses on transferring knowledge from a model trained on external unpaired data through weight sharing or weight composition. Due to the architecture of our model, we can simply "fake" visual concepts from out-of-domain textual corpus and pre-train our model on the faked concept-caption pairwise data.

%% file: content/Our_Model.tex

\section{Model}
\label{sec:model}

Our captioning system is illustrated in Fig.~\ref{fig:overview} and consists of the following steps. Given an input image we first use a Convolutional Neural Network to detect salient concepts which are then fed to a reviewer to output thought vectors. These vectors along with the groundtruth captions are then used to train a soft attentive decoder similar to the one proposed in~\cite{xu2015show}. We detail each step in the following sections.

\subsection{Visual Concept Detector}
The first step in our approach is to detect salient visual concepts in an image. We follow the approach proposed in~\cite{fang2015captions} and formulate this task as a multi-label classification problem. The set of output classes is defined by a dictionary $\V$ consisting of the 1000 most frequent words in the training captions, from which the most frequent 15 stop words were discarded. The set $\V$ covers around 92\% of the word occurrences in the training data.

As pointed out in~\cite{fang2015captions}, a naive approach to image captioning is to encode full images as feature vectors and use a multilayer perceptron to perform classification. However, most concepts are region-specific and~\cite{fang2015captions} demonstrated superior performance by applying a CNN to image regions and integrating the resulting information using a weakly-supervised variant of the Multiple Instance Learning (MIL) framework originally introduced in~\cite{maron1998framework}. We use this approach as the first step in our framework, we model the probability of a visual word $v_c \in \R^{|\V|}$ appearing in the image as a product of probabilities defined over a set of regions $\{ b_j \}$. Formally, we define this probability as
\begin{equation}
p(v_c) = 1 - \prod\nolimits_{b_j} \left( 1 - sigmoid(W^t_{v_c}\phi(b_{j}) + u_w)\right) 
\label{eq:MIL}
\end{equation}
where $j \in \{1, 2, \cdots, 14\times14\}$ indexes the image region from the response map $R_M(I)\in \R^{14 \times 14 \times A}$, $\phi(b_j) \in \R^{A}$ denotes the CNN features extracted over the region $b_j$, and $W^t_v \in \R^{|\V| \times A}$ and $u_w \in \R^{|\V|}$ are the parameters of the CNN learned from data. Our concept detector architecture is taken from~\cite{fang2015captions}, which is trained with maximum-likelihood on image-concept pairwise data extracted from MS-COCO. Note that the visual concept detector is trained only on MS-COCO and we use the same model on Flickr30k.

\begin{figure*}[t]
\begin{center}
\includegraphics[width=0.75\linewidth]{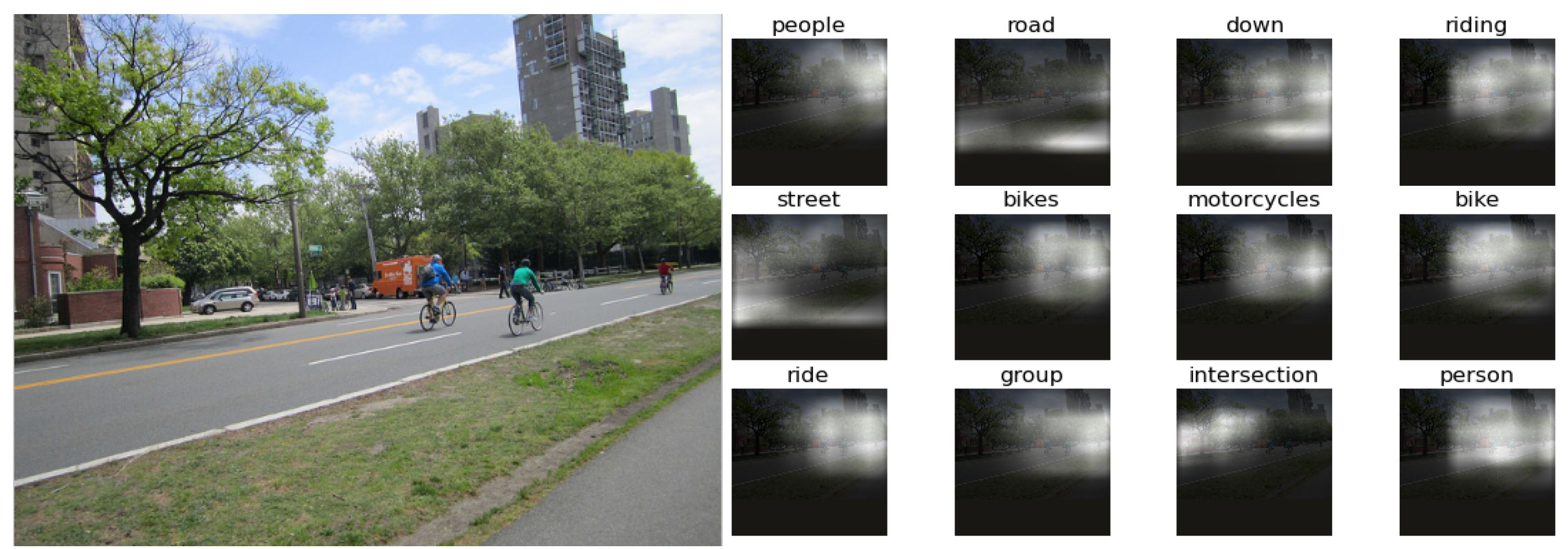}
\caption{\small{Attention of visual concepts. We select the top 12 words and visualize their attention weights in different regions of the image.}}
\label{fig:concept_attention}
\end{center}
\vspace{-1cm}
\end{figure*}

\subsection{Salient Regional Feature}
\label{sec:vis_feature}

A standard approach (see e.g.~\cite{you2016image}) to encode information about the image is by using the semantic word embeddings~\cite{pennington2014glove} corresponding to the detected visual concepts (we here consider the top $T$ concepts). The resulting word vectors are more compact than one-hot encoded vectors and capture many useful semantic concepts such as gender, location, comparative degrees, \dots

In this work, we also experiment with an approach that uses visual features extracted from the image sub-regions $B^{\tau}$ that have the strongest connections with the set of detected visual concepts $v_c^{\tau}$. For each of the top $T$ concepts in $\{ v^{\tau}_c \}_{\tau=1}^T$, we compute image sub-regions $\{B^{\tau}\}^T_{\tau=1}$ as
\begin{equation}
B^{\tau} = \argmax{b_j} W^t_{v^{\tau}_c} \phi(b_j).
\label{eq:spatial-info}
\end{equation}
In summary, we extract salient features$\{ v^{\tau}_c \}_{\tau=1}^T$ from an image either in two ways:
\begin{equation}
v_{\tau} = \begin{cases}
E v^{\tau}_c & semantic\\ 
\phi(B^{\tau}) & visual\\
\end{cases}
\label{eq:embedding}
\end{equation}
where $E$ is an embedding matrix that maps a one-hot encoded vector $v_c \in \R^{|\V|}$ to a more compact embedding space $v_{\tau} \in \R^d$, and $\phi(B^{\tau}) \in \R^{A}$ corresponds to the CNN features from image region $B^{\tau}$. Note that $v_{\tau}$ has a different dimension in the two cases, as semantic feature $d=300$ while as visual feature $d=A=2048$.

As demonstrated in Section~\ref{sec:experiments}, we found that using CNN regional features can be advantageous over semantic word features. Our conjecture is that visual features are often more expressive since one image region can relate to multiple word choices.

\subsection{Input Reviewer}
\label{sec:inp_review}

As depicted in~\autoref{fig:concept_attention}, most of the detected visual concepts tend to capture very salient image concepts. However, some concepts are duplicated and others are incorrect such as "intersection" in the example provided in~\autoref{fig:concept_attention}. We address this problem by applying an input reviewer~\cite{yang2016encode} to synthesize "thought vectors" that capture globally consistent concepts and eliminate mistakes introduced by the visual concept detector.
Note that unlike the approach described in~\cite{yang2016encode} that takes serialized CNN features for the whole image as input, we instead use the features described in Section~\ref{sec:vis_feature}. Since these features already have a strong semantic meaning, we did not apply the "weight typing" and "discriminative supervision" strategies suggested in~\cite{yang2016encode}.

Our input reviewer is composed of an attentive LSTM, which estimates an attention weight $\beta_{\tau, t}$ for a given $v_{\tau}$ and outputs its hidden state as thought vectors $ f_t \in \R^{F}$. Formally,
\begin{equation}
\beta_{t, \tau} = \frac{\exp(g_e(f_{t-1}, v'_{t-1}, v_{\tau}))}{\sum_{\tau} \exp(g_e(f_{t-1},  v'_{t-1}, v_{\tau}))},
\label{eq: input-beta}
\end{equation}
where $v'_{t-1}$ is the overview context vector from the last step. We use an attention function $g_e$ with parameters $W^e$, defined as
\begin{equation}
g_e(f_{t-1}, v'_{t-1}, v_{\tau}) =
        {W^e_a}^T \tanh(v_{\tau} + W^e_f [f_{t-1};v'_{t-1}])
\label{eq:attention_mechanism_encoder}
\end{equation}
where $W^e_f \in \R^{d \times (d+F)}$ and $W^e_a \in \R^{d}$ are parameters learned from data, we set $F=300$ in our experiment.

In order to select which visual concepts to focus on, we could sample with regard to the attention weights $\beta$, but a simpler approach described in~\cite{bahdanau2014neural} is to take a weighted sum over all inputs, i.e.
\begin{equation}
v'_{t} = \sum\nolimits_{\tau} \beta_{t,\tau} v_{\tau}.
\label{eq: soft-sum}
\end{equation}

\begin{figure}[ht]
\begin{center}
   \includegraphics[width=0.7\linewidth]{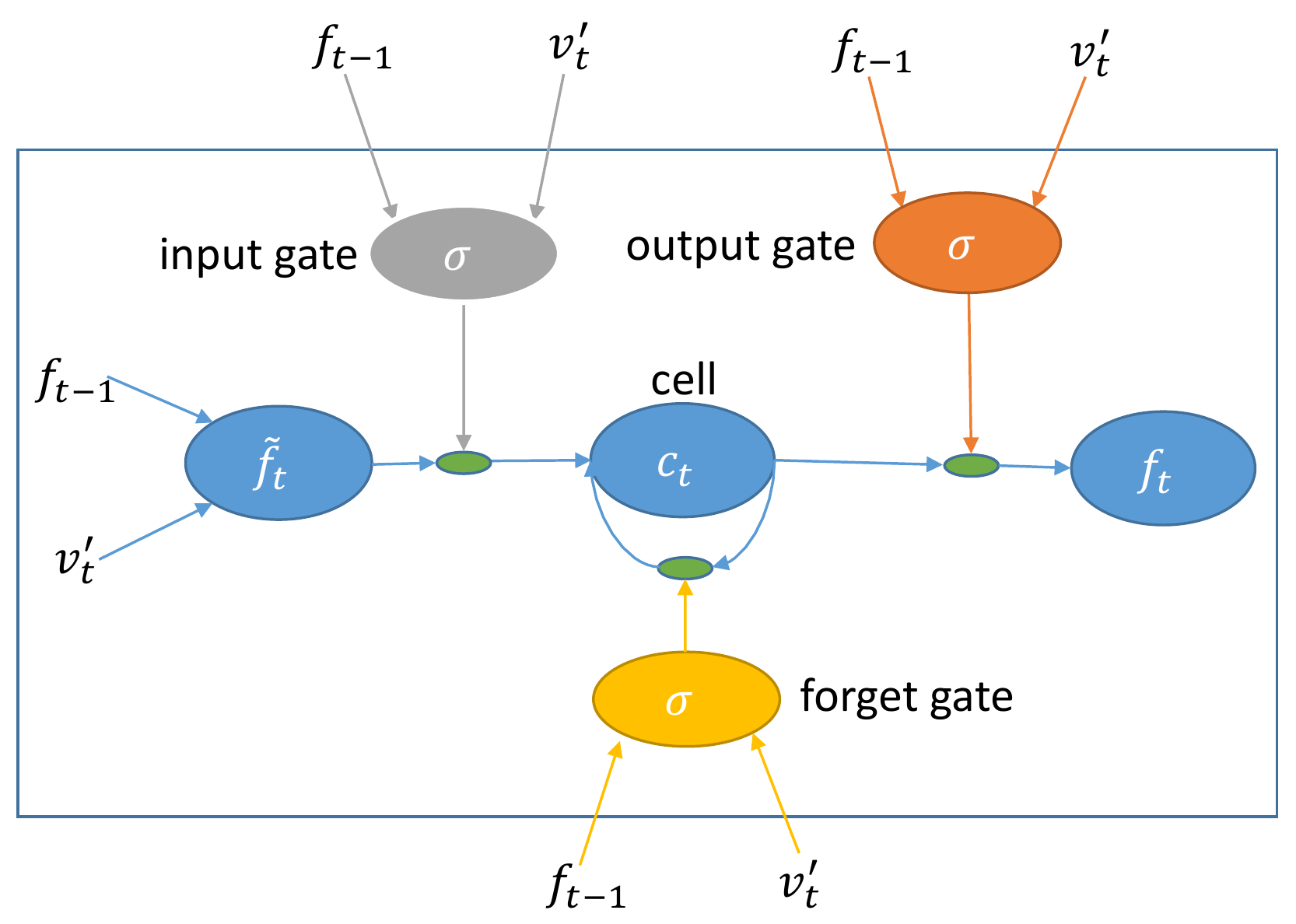} \\
   \includegraphics[width=0.7\linewidth]{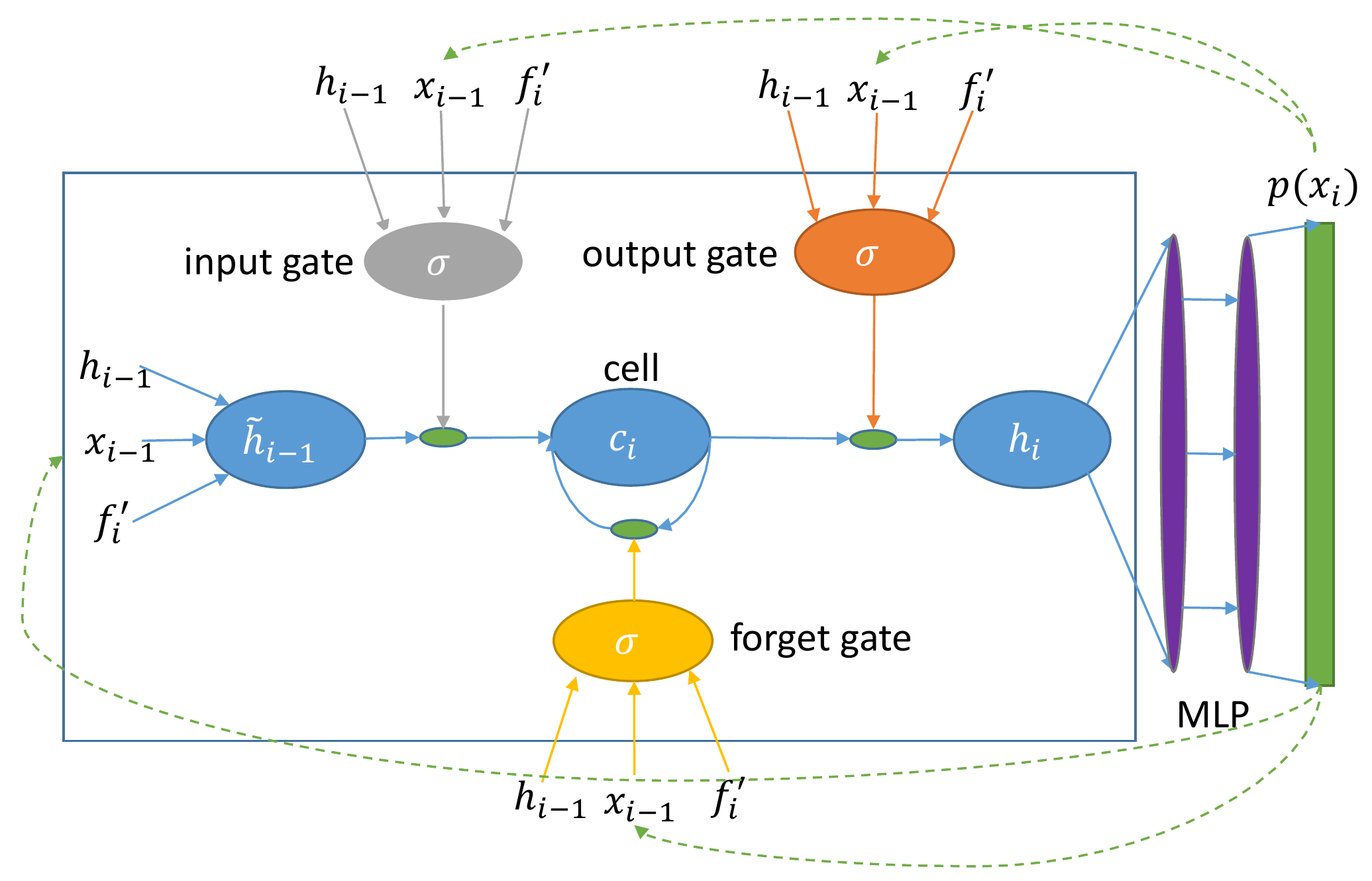}
   \caption{\small{LSTM model for the input reviewer (top) and the decoder (bottom).}}
   \label{fig:review-decode-lstm}
\end{center}
\vspace{-1cm}
\end{figure}

As shown on~\autoref{fig:review-decode-lstm}, our LSTM reviewer uses $v'_t$ and $f_{t-1}$ as inputs to produce the next thought vector $f_t$. Unlike the LSTM decoder presented in the next section, it does not rely on the input symbols $\{x_i\}$. The reviewer LSTM basically functions as a text synthesizer without any reliance on visual contexts, which explains why we can pre-train this part using only textual data (see Section~\ref{sec:semi_supervised_training}).

\subsection{Soft Attentive Decoder}

Our decoder is based on~\cite{xu2015show} and is formulated as an LSTM network with attention mechanism on the thought vectors. The decoder LSTM depicted in \autoref{fig:review-decode-lstm} takes as input both the set $\{f_t \}$ and input symbols $\{x_i\}$ from groundtruth captions (or word predictions that approximate the word distribution). The decoder estimates an attention weight $\alpha_{i,t}$ based on the past context overview vector $c_{i-1}$, past hidden state $h_{i-1} \in \R^H$ and thought vectors $\{f_t\}$. 
Formally, we write it as
\begin{flalign}
\alpha_{i, t} &= \frac{\exp(g_d(f'_{i-1}, h_{i-1}, f_t))}{\sum_{t} \exp(g_d(f'_{i-1}, h_{i-1}, f_t))}, \\
g_d(f'_{i-1}, h_{i-1}, f_t) &=
        {W^d_a}^T \tanh(f_t + W^d_f [h_{i-1};f'_{i-1}]), \nonumber
\label{eq:attention_mechanism_decoder}
\end{flalign}
where $W^d_f \in \R^{F \times (F+H)}$, and $W^d_a \in R^{F}$, we use $H = 1000$ in our experiments.

Similarly to the input reviewer, we use a weighted sum $f'_i$ over all thought vectors to approximate sampling
\begin{equation}
f'_i = \sum\nolimits_t \alpha_{i,t} f_t.
\label{eq:soft_attention}
\end{equation}

Unlike the input reviewer whose initial state is set to zero, we introduce visual information in the decoder by initializing the LSTM memory $c_0$ and state $h_0$ with a linear transformation of CNN features, i.e.
\begin{align}
h_0 = W_h \psi(I), \quad c_0 = W_m \psi(I),
\label{eq: init_state_mem}
\end{align}
where $\psi(I)$ denotes the CNN features of the image, $W_h \in \R^{H \times |\psi(I)|}$ and $W_m \in \R^{H \times |\psi(I)|}$ are parameters learned from data. Note that $\psi(I)$ is different from $\phi(B^{\tau})$ in~\autoref{eq:MIL} in that $\psi(I)$ extracts full image features from the upper layer, while $\phi(B^{\tau})$ extracts sub-region features from the response map. 

\subsection{Model Learning}
The output state $h_i$ of the decoder LSTM contains all the useful information for predicting next word $x_i$. We follow the implementation of~\cite{xu2015show} and add a two-layer perceptron with dropout on top of the decoder LSTM to predict the distribution for all words in the vocabulary. We calculate the cross-entropy loss based on the proposed distribution $p(x_i)$ and groundtruth word $y_i$.

We train our model using maximum likelihood with a regularization term on the attention weights $\alpha$ and $\beta$ of the input reviewer and attentive decoder.
Formally, we write
\begin{equation}
Loss = -\min_{\theta} \sum\nolimits_i \log{p(y_i)} + \lambda (g(\alpha) + g(\beta))
\end{equation}
\begin{equation}
g(\alpha) = \sum\nolimits_t(1 - \sum\nolimits_i \alpha_{i,t})^2,
\end{equation}
where $y_i$ is the groundtruth word, $\theta$ refers to all model parameters and $\lambda > 0$ is a balancing factor between the cross-entropy loss and a penalty on the attention weights. We use the penalty function $g$ described in~\cite{xu2015show} to ensure every concept and thought vector receives enough attention.  

\subsection{Semi-supervised Learning}
\label{sec:semi_supervised_training}

\begin{figure*}[ht]
\begin{center}
   \includegraphics[width=0.8\linewidth]{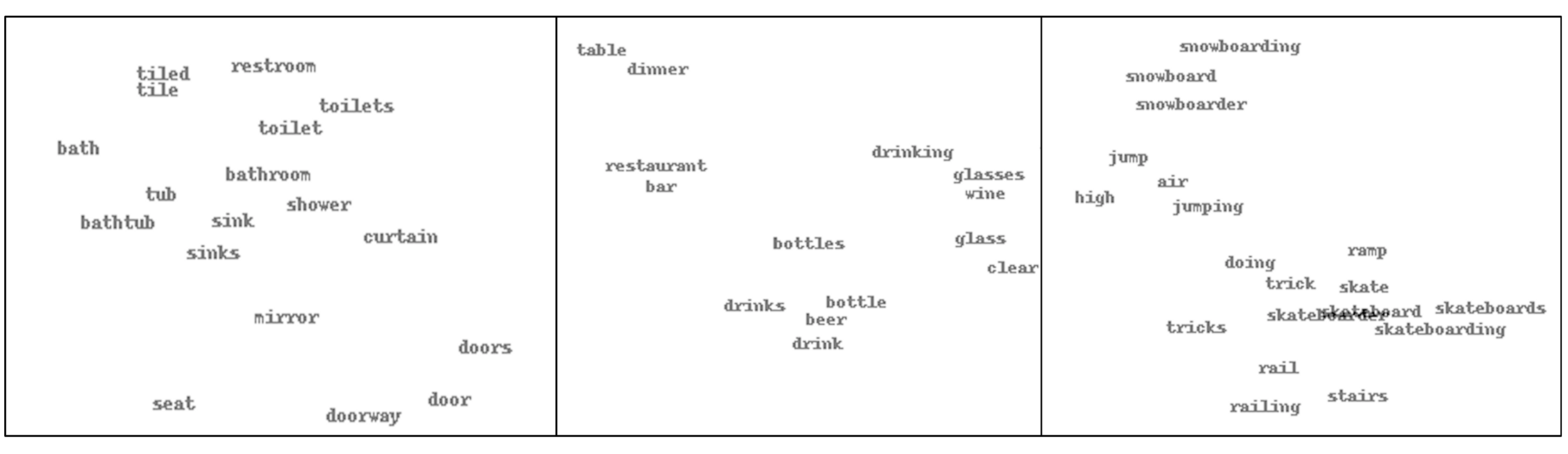}
   \caption{\small{\textit{Visualization of the faked visual regional features.} We here show a projection of the features obtained by t-SNE for three different regions of the feature space. Words that are semantically or morphologically similar are clearly clustered together.}}
   \label{fig:regional_cluster}
   \vspace{-5mm}
\end{center}
\end{figure*}

Most existing approaches to image captioning rely on pairs of images and corresponding captions for learning their model parameters. Such training data is typically expensive to produce and usually requires using crowd-sourcing techniques. The MS-COCO dataset was for instance annotated using Amazon's Mechanical Turk, a process that required $70K$ worker hours~\cite{lin2014microsoft}. In contrast, unpaired text and image data is abundant and cheap to obtain but can not be used as is with current image captioning systems. We here suggest a novel approach to exploit text data without corresponding images to train our model. Since images are used as inputs to the visual concept detector to generate visual concepts, we need to "fake" such information during the unsupervised training phase. We here propose two different methods for each of the two possible ways to encode visual concepts.

\begin{figure}[t]
\begin{center}
\fbox{\rule{0pt}{0in} 
   \includegraphics[width=0.95\linewidth]{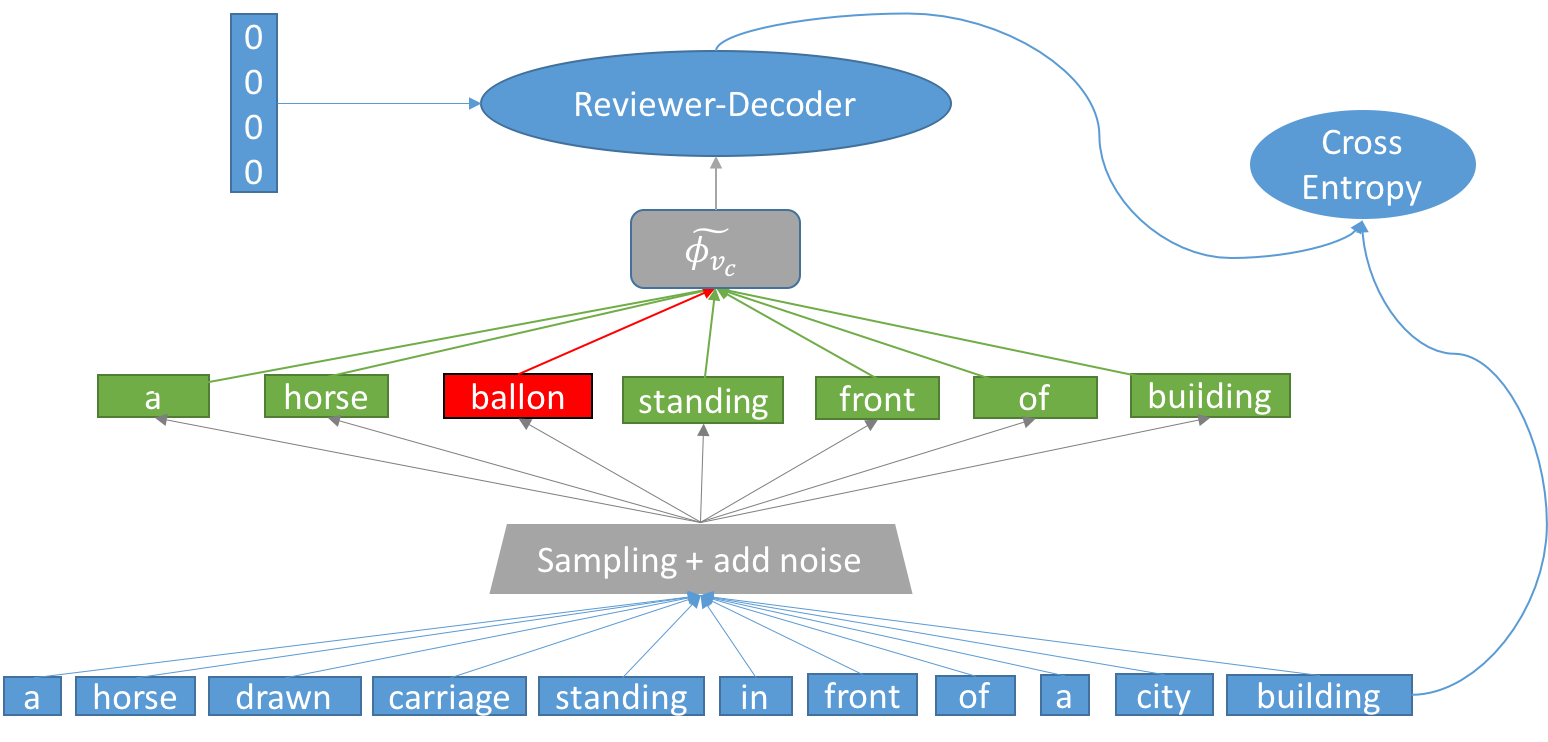}
}
\end{center}
   \caption{\small{\textit{Faking Semantic Embeddings}. We here illustrate how we train our model using out-of-domain text data. Starting from a sentence without corresponding annotations (blue boxes), we sample a given number of concepts shown in the green boxes. The red box shown in the example above is a noisy concept artificially introduced to make the model more robust.}}
\label{fig:pre-training}
\vspace{-5mm}
\end{figure}

\paragraph{Fake Semantic Embeddings}
In the case where the salient features$\{ v^{\tau}_c \}_{\tau=1}^T$ described in Section~\ref{sec:vis_feature} are based on semantic embeddings, we can directly fake these concepts based on the groundtruth sentences. This process is illustrated
in~\autoref{fig:pre-training}. We experiment with two methods named "Truth Generator" and "Noisy Generator". The "Truth Generator" approach takes sample words from sentences longer than 15 words or zero-pad shorter sentences to generate 15 concepts. The "Noisy Generator" mixes words sampled from the groundtruth sentences with randomly sampled words to form 15 concepts. Further details are provided in the appendix. Besides, we also experimented with out-of-domain text data with different sizes, i.e. 600K and 1.1M captions, which roughly corresponds to the number of training captions in MS-COCO.

\paragraph{Fake Regional Visual Features}
The case of using salient regional features is more difficult to handle since our additional training data only consists of textual data without corresponding images. We propose to address this problem by relying on the strong correlation between visual concept and regional features. Specifically we construct a mapping $\bar{\phi}$ from the concept space to the regional feature space. For simplicity, we assume the regional feature $\phi$ corresponding to each concept $v_c$ follows a gaussian distribution. Thus, we can estimate its mean value by averaging all the regional features associated given concept in the training data, i.e.
\begin{equation}
\bar{\phi}(v_c) = \mathbb{E}_{\tau': v_c^{\tau'}=v_c}[\phi(B^{\tau'})]
\end{equation}
where $\bar{\phi} \in \R^{2048}$. We visualized these "faked" regional features using t-SNE~\cite{maaten2008visualizing} and the results shown in~\autoref{fig:regional_cluster} demonstrate that the aggregated regional features capture similar properties to the ones of the semantic embeddings. Once we have established such mapping, we can artificially encode a given text as regional features. The ``faked'' regional features can thus be used as inputs for the unsupervised learning phase.

\paragraph{Unsupervised training}
This training phase results in a two-step procedure. The first step is to pre-train our model on unpaired textual data, which teaches the model to produce captions based on out-of-domain language samples. Note that more than 60\% of all the parameters can be pre-trained with our unsupervised learning framework, except the transformation matrix $W_h, W_m$ used to initialize $h_0$, $c_0$ with CNN features (further details are given in the appendix). In the second phase of training, we optimize our model on in-domain supervised data (i.e. pairwise MS-COCO image-text dataset). As can be seen from~\autoref{fig:learning curve}, after only one epoch of in-domain adaptation, the performance already reaches a quite promising stage. 

%% file: content/Experiments.tex

\section{Experiments}
\label{sec:experiments}

\subsection{Data}
We evaluate the performance of our model on the MS-COCO~\cite{lin2014microsoft} and Flickr30K\cite{plummer2015flickr30k} datasets. The MS-COCO dataset contains 123,287 images for training and validation and 40775 images for testing, while Flickr30K provides 31,783 images for training and testing. For MS-COCO, we use the standard split described by Karpathy~\footnote{\url{https://github.com/karpathy/neuraltalk2}} for which 5000 images were used for both validation and testing and the rest for training. For Flickr30K, we follow the split of~\cite{jin2015aligning} using 1K images for both validation and test and the rest for training. During the pre-training phase, we use both the 2008-2010 News-CommonCrawl and Europarl corpus~\footnote{\url{http://www.statmt.org/wmt11/translation-task.html\#download}} as out-of-domain training data. Combined, these two datasets contain $\sim3M$ sentences, from which we removed sentences shorter than 7 words or longer than 30 words. We also filter out sentences with unseen words in the MS-COCO dataset. After filtering, we create two separate datasets of size 600K and 1.1M, which are then both tokenized and lowercased, and used for the pre-training phase. We train the model with a batch size of 256 and validate on an out-of-domain held-out set. The training is ended when the validation score converges or the maximum number of epochs is reached. After pre-training, we then use the trained parameters to initialize the in-domain training stage.

\subsection{Experimental Setting}
We use GloVe~\cite{pennington2014glove}~\footnote{~\url{https://github.com/stanfordnlp/GloVe}} to train 300-dimensional semantic word embeddings on Wiki+Gigaword. We use full image features extracted from the CNN architecture, as in~\cite{simonyan2014very,he2015deep}, to initialize the decoder LSTM. In our experiments, we set the batch size to 256, vocabulary size to 9.6K, reviewer LSTM hidden size to 300 and decoder LSTM hidden layer size to 1000. We use rmsprop~\cite{tieleman2012lecture} with a learning rate of $10^{-4}$ to optimize the model parameters. Training on MS-COCO takes around $1$ day to reach the best performance. We do model selection by evaluating the model on the validation set after every epoch, with the maximum training epoch set to 20. We keep the model with the best BLEU-4 score and evaluate its performance on the test set. We here only report the model performance on the test set. At test time, we do beam search with a beam size of 4 to decode words until the end of sentence symbol is reached.

\subsection{Evaluation Results}
\input{content/Result_table_unofficial}

We use different standard evaluation metrics described in~\cite{chen2015microsoft}, including BLEU~\cite{papineni2002bleu}, a precision-based machine translation evaluation metric, METEOR~\cite{banerjee2005meteor}, as well as CIDEr \cite{vedantam2015cider} which is a measure of human consensus.
The results are shown in~\autoref{tab:result} where "Ours-x" indicates the performance of different variants of our model. The "Baseline" model takes visual concept embeddings as inputs to the attentive decoder without using any pre-training or visual feature for initialization. The "Rev" variant adds an input reviewer in front of the attentive decoder to synthesize salient features from the images. The models with "Fc7" and "Pool5" respectively use the fc7 layer from VGG~\cite{simonyan2014very} and Pool5 layer from ResNet152~\cite{he2015deep} for the decoder initialization. The latter brings significant improvements across all metrics. Models with "Sm" use semantic embeddings as input to the reviewer, while "Rf" use regional features. Models with "Bsl" use our pre-trainig method while "large" corresponds to using the 1.1M corpus, "small" is for the 660K corpus, and "noisy" means applying the Noisy Generator. Finally, "Ens" means using an ensemble strategy to combine the results of 5 identical models "Ours-Pools5-Rev-Rf-Bsl" which were trained independently with different initial parameters. 

\begin{figure}[htb]
\begin{center}
\includegraphics[width=0.8\linewidth]{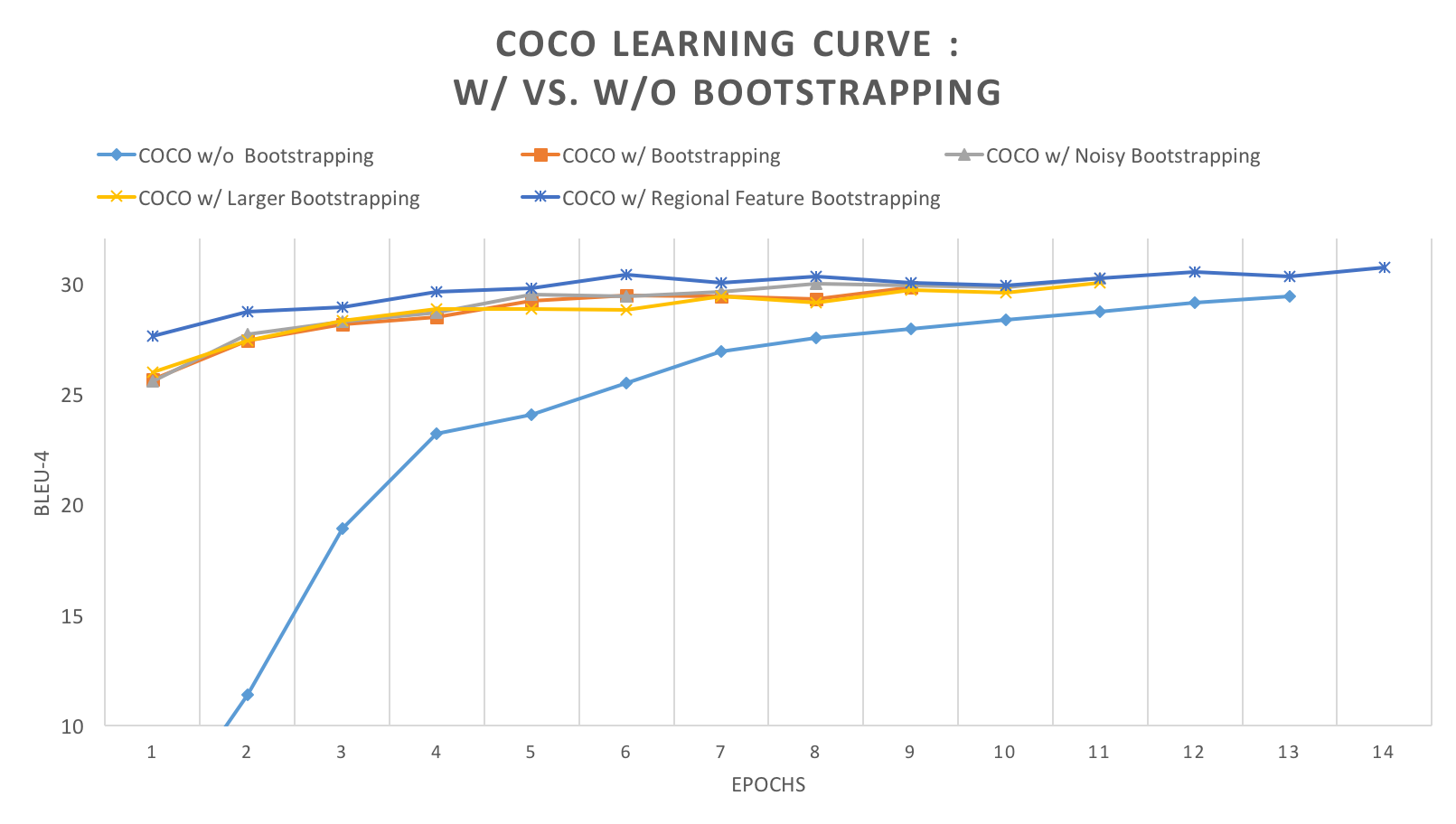}
\includegraphics[width=0.8\linewidth]{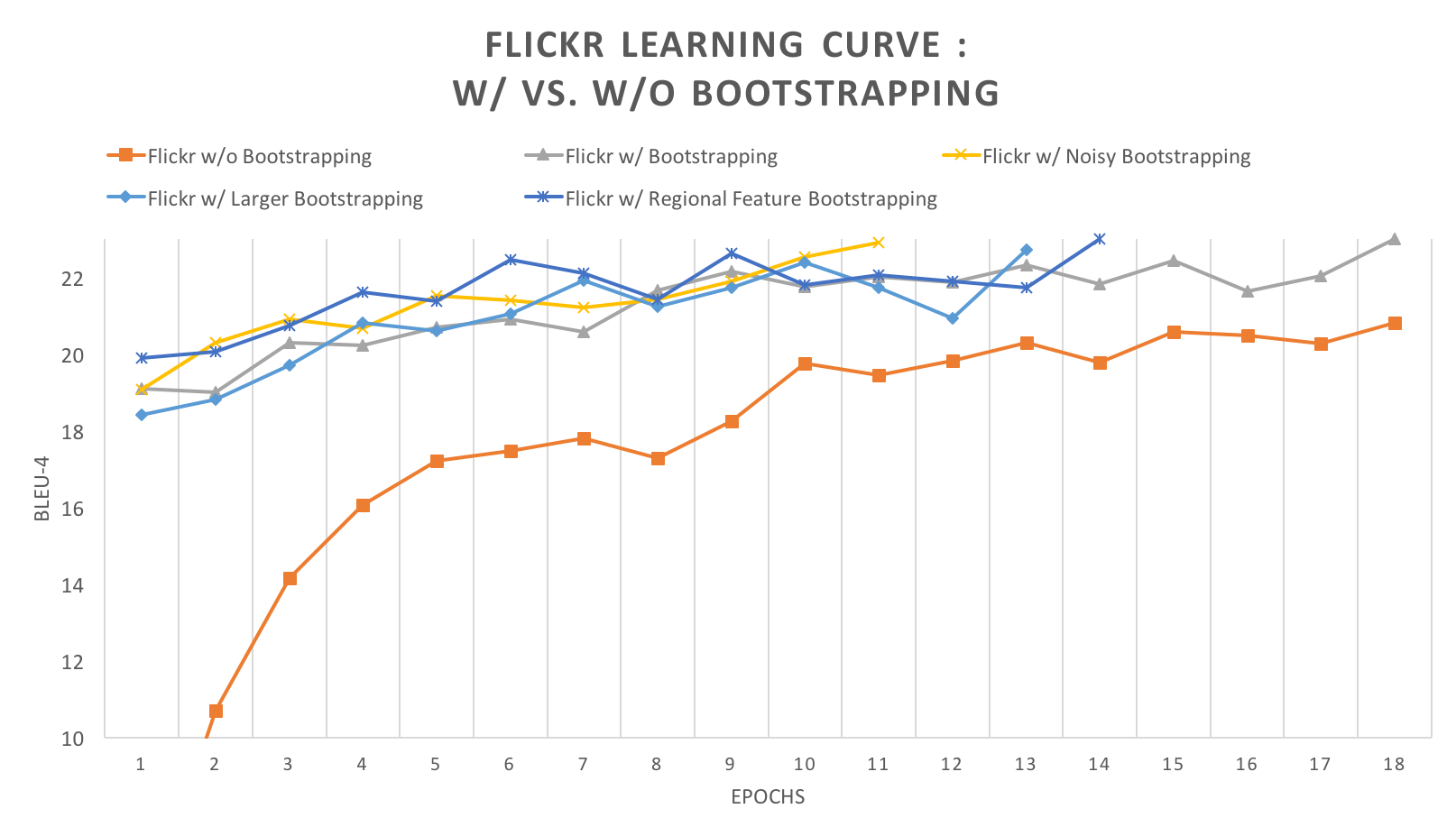}
\end{center}
   \caption{\small{Analysis of the impact of the unsupervised learning phase on training time. The y axis represents the BLEU-4 score on the validation set and the x axis denotes the number of epochs.}}
\label{fig:learning curve}
\vspace{-1cm}
\end{figure}

Our model without the unsupervised learning phase (Ours-Fc7-Rev-Sm) gets similar performance to ERD+VGG~\cite{yang2016encode}. When pre-training with out-of-domain data, our model outperforms its rival consistently across different metrics. We have also observed that the improvements on Flickr30K is more significant than on MS-COCO, which might partly be due to the smaller amount of training data for Flickr30K. When pre-training with out-of-domain data and combined with the reviewer module and ResNet152 features, our ensemble model improves significantly across several metrics and achieves state-of-art performance on both datasets. This clearly demonstrates that the unsupervised learning phase can not only increase n-grams precision but also adapts to human consensus by generating captions that are more diverse.
\paragraph{Semantic Embedding vs. Regional Features} The results in~\autoref{tab:result} show that regional features yield higher scores for most  metrics. We also report results for an ensemble combining both features (see ours-Pool5-Rev-Rf-Sm-Bsl-Ens), which is shown to be inferior to the ensemble based on regional features alone (see Ours-Pool5-Rev-Rf-Bsl-Ens).
\input{content/Result_table}
\paragraph{Semi-supervised Learning.} We show the evolution of the BLEU-4 score on the validation set in~\autoref{fig:learning curve}. We can see that when pre-trained on unsupervised data, the model starts with a very good accuracy and keeps increasing afterwards. In the end it outperforms the non pre-trained model by a large margin. We also experimented with the "Truth Generator" and "Noisy Generator" variants described in Section~\ref{sec:semi_supervised_training} with varying size of the corpus. The results are shown in~\autoref{tab:result}. We observe that adding noise improves the performance in terms of most metrics, which indicates that a model trained with additional noise is more robust, thus producing more accurate captions. Besides, we see that simply enlarging the size of the training corpus (model with "large" in the title) does not help achieve significantly better scores, which might be due to the fact that the additional data is taken from the same source as the smaller one.
\begin{figure}[ht]
\begin{center}
\includegraphics[width=1.0\linewidth]{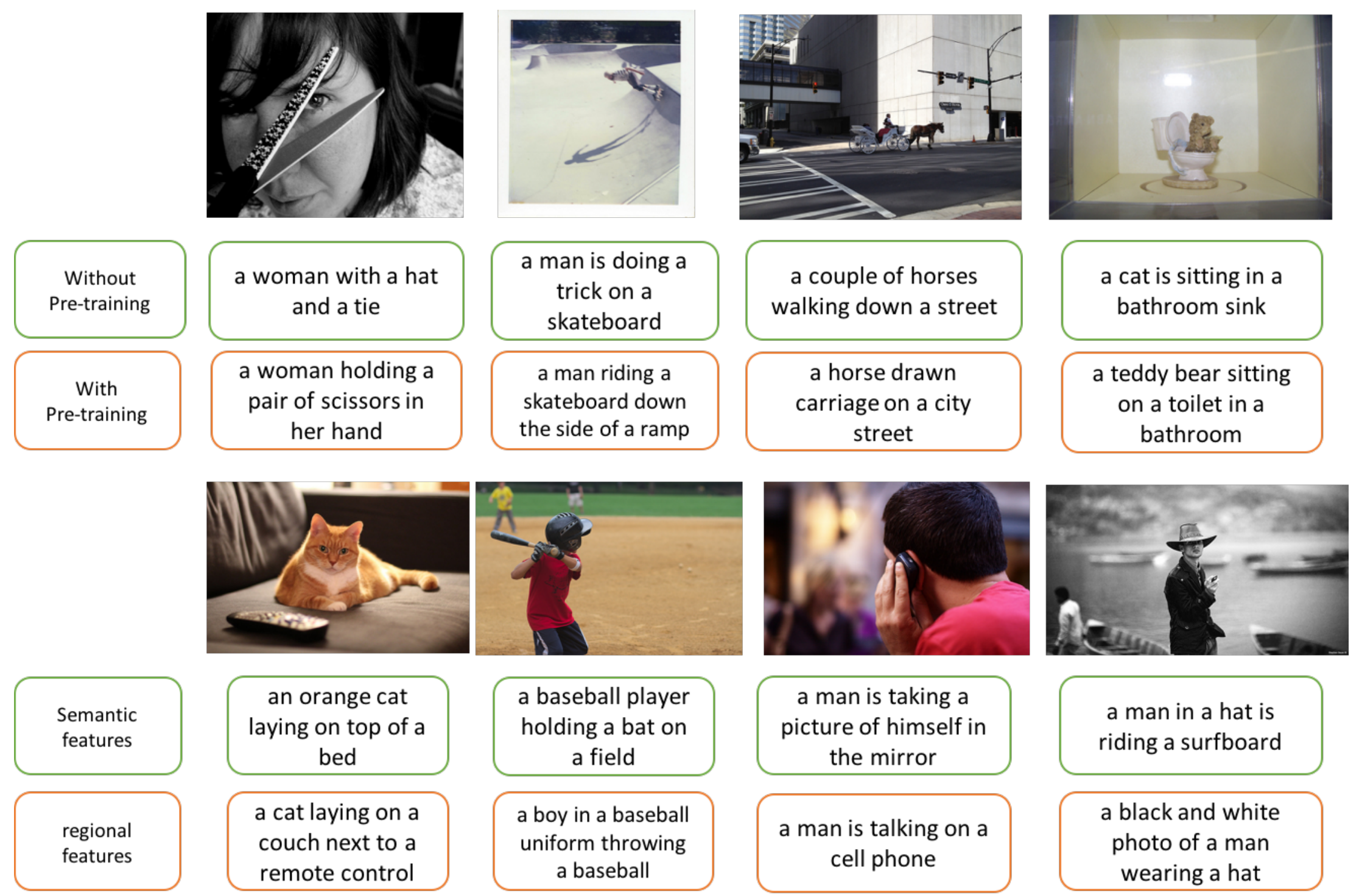}
\end{center}
   \caption{\small{Qualitative analysis of the impact of the pre-training procedure as well as the use of visual regional features.}}
\label{fig:example_images}
\vspace{-1cm}
\end{figure}
\paragraph{Sample Results.} We show examples of the captions produced by our model in~\autoref{fig:example_images}. We would like to make two observations from these examples: (1) using a pre-training phase on additional out-of-domain text data yields a model that can produce a wider variety of captions and (2) the regional features captures more adequate visual concepts which then yields more accurate textual descriptions.
\paragraph{Results on MS-COCO testing server}
We also submitted our results to the MS-COCO testing server to evaluate the performance of our model on the official test set. \autoref{tab:official_result} shows the performance Leaderboard for 5 reference captions (c5) and 40 reference captions (c40). Note that we applied the same setting as the best model reported in~\autoref{tab:result}. Our model ranks among the top 10 in the Leaderboard.

%% file: content/Result_table_unofficial.tex
\begin{table*}[ht]
\centering
\begin{tabular}{|l|l|l|l|l|l|l|l|l|l|l|}
\hline
\multirow{2}{*}{\backslashbox{Dataset}{Model}} & \multicolumn{6}{c|}{MS-COCO}  & \multicolumn{4}{c|}{Flickr30K}\\
\cline{2-11}
& B-1 & B-2 & B-3 & B-4 & METEOR & CIDEr &   B-1 & B-2 & B-3 & B-4 \\\hline
Neuraltalk2 \cite{karpathy2015deep} & 62.5 & 45.0 & 32.1 & 23.0 & 19.5 & 66.0 & 57.3 & 36.9 & 24.0 & 15.7 \\\hline
Soft Attention \cite{xu2015show} & 70.7 & 49.2 & 34.4 & 24.3 & 23.9 & - & 66.7 & 43.4 & 28.8 & 19.1 \\\hline
Hard Attention \cite{xu2015show} & 71.8 & 50.4 & 35.7 & 25.0 & 23.0 & - & \secondary{66.9} & 43.9 & 29.6 & 19.9 \\\hline 
MSR \cite{fang2015captions} & - & - & -  & 25.7 & 23.6 & - & - & - & - & - \\\hline 
Google NIC \cite{vinyals2015show} & 66.6 & 46.1 & 32.9 & 24.6 & - & - & 66.3 & 42.3 & 27.7 & 18.3 \\\hline
TWS \cite{mao2015learning} & 68.5 & 51.2 & 37.6 & 27.9  & 22.9 & 81.9 & - & - & - & - \\\hline 
ATT-FCN(Ens) \cite{you2016image} & 70.9 & 53.7 & 40.3 & 30.4 & 24.3 & - & 64.7 & 46.0 & 32.4 & 23.0 \\\hline
ATT-FCN(Sin) \cite{you2016image} & 70.4 & 53.1 & 39.4 & 29.3 & 23.9 & - & 62.8 & 43.7 & 30.1 & 20.7\\\hline
ERD+VGG \cite{yang2016encode} & - & - & - & 29.0 & 24.0 & 89.5 & - & - & - & -\\
\hline
\hline
Ours-Baseline & 68.2 & 50.7 & 37.1&  26.7 & 23.4 & 84.2 & 61.8 & 41.9 & 28.2 & 18.8\\\hline
Ours-Fc7-Sm & 68.6 & 50.7 & 37.3 & 27.7 & 23.6 & 85.5 &61.9 & 43.0& 29.4&19.6 \\\hline
Ours-Fc7-Rev-Sm & 70.2 & 53.3 & 39.3 & 28.8 & 23.4 & 87.8 &62.5 & 43.0 & 29.1& 19.7\\\hline
Ours-Fc7-Rev-Sm-Bsl(small) & 70.1 &53.9 &39.9 & 29.5&  23.8&  90.4 & 63.5 & 44.3 & 30.5 & 20.8 \\\hline
Ours-Pool5-Rev-Sm-Bsl(small) & 72.2 & 54.6 & 40.4 & 29.8 & 24.3& 92.7& 66.1 & 47.2 & 33.1 & 23.0 \\\hline
Ours-Pool5-Rev-Sm-Bsl(large) & 72.3 & 54.7 & 40.5 & 30.0 & 24.5& 93.4 & 66.5 & 47.3 & 33.1 & 22.7 \\\hline
Ours-Pool5-Rev-Sm-Bsl(noisy) & 72.6 & 55.0 & 40.8 & 30.2 & 24.7 & 94.0 & 66.6 & 47.3 & 33.2 & 22.9 \\\hline
Ours-Fc7-Rev-Rf & 70.6 & 53.6 & 39.5 & 29.0 & 23.6 & 87.4 & 61.8 & 42.9 & 29.4 & 20.0\\\hline
Ours-Fc7-Rev-Rf-Bsl & 71.4 & 54.6 & 40.6 & 30.1 & 24.3 & 91.3 & 64.2 & 45.5 & 31.7 & 21.9 \\\hline
Ours-Pool5-Rev-Rf-Bsl & 72.5 & 55.1 & 41.0 & 30.6 & 24.8 & 95.0 & 66.4 & 47.3 & 33.3 & 23.0 \\\hline
Ours-Pool5-Rev-Rf-Sm-Bsl-Ens & \secondary{72.9} & \secondary{55.8} & \secondary{41.6} & \secondary{30.9} & \secondary{24.8} & \secondary{95.8} & \secondary{66.9} & \secondary{47.8} & \secondary{33.7} & \secondary{23.3} \\\hline
Ours-Pool5-Rev-Rf-Bsl-Ens & \highlight{73.4} & \highlight{56.5} & \highlight{42.5} & \highlight{32.0} & \highlight{25.2} & \highlight{98.2} & \highlight{67.2} & \highlight{48.2} & \highlight{34.0} & \highlight{23.8} \\\hline
\end{tabular}

\caption{\small{Performance in terms of BLEU-1,2,3,4, METEOR and CIDEr compared to other state-of-the-art methods on MS-COCO and Flickr30K dataset. For the competing methods, we report the performance results cited in the corresponding references. The numbers in red denotes the best known results, the numbers in blue denotes the second best known results, (-) indicates unknown scores. Note that all the scores are reported in percentage.}}
\label{tab:result}
\vspace{-5mm}
\end{table*}

%% file: content/Result_table.tex
\begin{table*}[htb]
\begin{center}
\begin{tabular}{|l|l|l|l|l|l|l|l|l|l|l|l|l|}
\hline
\multirow{2}{*}{\backslashbox{Dataset}{Model}}  & \multicolumn{2}{c|}{B@1} & \multicolumn{2}{c|}{B@2} & \multicolumn{2}{c|}{B@3} & \multicolumn{2}{c|}{B@4} & \multicolumn{2}{c|}{CIDEr} & \multicolumn{2}{c|}{METEOR} \\ \cline{2-13}
 & c5 & c40 & c5 & c40 & c5 & c40 & c5 & c40 & c5 & c40 & c5 & c40 \\ \hline
ATT-LSTM-EXT (Ours) & 73.4 & 91.0 & 56.3 & 82.3 & 42.3 & 71.4 & 31.7 & 60.2 & 96.4 & 97.4 & 25.4 & 34.1 \\ \hline
ATT~\cite{you2016image} & 73.1 & 90.0 & 56.5 & 81.5 & 42.4 & 70.9 & 31.6 & 59.9 & 94.3 & 95.8 & 25.0 & 33.5 \\ \hline
Google~\cite{vinyals2015show} & 71.3 & 89.5 & 54.2 & 80.2 & 40.7 & 69.4 & 30.9 & 58.7 & 94.3 & 94.6 & 25.4 & 34.6 \\ \hline
kimiyoung~\cite{yang2016encode} & 72.0 & 90.0 & 55.0 & 81.2 & 41.4 & 70.5 & 31.3 & 59.7 & 96.5 & 96.9 & 25.6 & 37.7 \\ \hline
\end{tabular}
\caption{\small{Leaderboard of the published state-of-the-art image captioning models on the online COCO testing server (http://mscoco.org/dataset/\#captions-leaderboard), where B@N, M, R, and C are short for BLEU@N, METEOR, and CIDEr scores. All values are reported as percentages (\%).}}
\label{tab:official_result}
\end{center}
\vspace{-5mm}
\end{table*}

%% file: content/Conclusion.tex

\section{Conclusion}
We proposed a novel training method that exploits external text data without requiring corresponding images. This yields significant improvements in terms of the ability of the language model to generate more accurate captions. We also introduced a new model that brings some new improvements such as using salient regional features instead of traditional semantic word embeddings. Our new model together with the suggested pre-training method achieves state-of-the-art performance. Given the wide availability of text data, our pre-training method has the potential of largely improving the generality of most existing image captioning system, especially for domains with little paired training data.

%% file: content/Appendix.tex
\section{Acknowledgments}

We thank Prof. Juergen Rossmann and Dr. Schristian Schlette from RWTH Aachen University for their support as well as the Aachen Super Computing Center for providing GPU computing.

\appendix

\section{Appendix - Implementation details}

\subsection{Input Reviewer}
The input reviewer uses an LSTM to generate thought vectors. Formally, a thought vector $f_t$ is computed as 
\begin{align}
f^e_t &= \sigma(W^e_r [f_{t-1}, v'_t] + b^e_r) \\
o^e_t &= \sigma(W^e_o [f_{t-1}, v'_t] + b^e_i) \\
i^e_t &= \sigma(W^e_i [f_{t-1}, v'_t] + b^e_o) \\
\widetilde{f}_t &= \tanh(W^e_h [f_{t-1}, v'_t] + b^e_h) \\
c^e_t &= f^e_t c^e_{t-1} + i^e_t \widetilde{f}_t \\
f_t &= o^e_t \tanh(c^e_t),
\end{align}
where $f^e_t, o^e_t, i^e_t, \widetilde{f}_t, c^e_t, f_t  \in \R^{F}$ are the forget/output/input gates and cell input/hidden/output states. These gates and states are controlled by the last thought vector $f_{t-1}$ and overview feature vector $v'_t$. $W^e_r, W^e_o, W^e_i, W^e_h \in \R^{F \times (F + d)}$, $b^e_r, b^e_o, b^e_i, b^e_h \in \R^{F}$ are the LSTM parameters learned from data. We use the LSTM cell output states $\{f_t\}_{t=1}^T$ directly as thought vectors. We set the LSTM state size $F=300$ in our experiments.

\paragraph{Semantic Features}
When semantic features are used as input to the reviewer, we set $d=300$, which is the same as the GloVe embedding size. As described, the reviewer thus contains around $F \times (F + d) \times 4 = 0.72M$ parameters.

\paragraph{Regional Features}
When regional features are used as input to the reviewer, we set $d=2048$, which corresponds to the dimension of the convolutional feature map of the visual concept detector. The reviewer thus contains around $F \times (F+d) \times 4 = 2.8M$ parameters. 

\subsection{Decoder}
Our decoder is also based on an LSTM architecture, but unlike the reviewer described previously, it also involves the embedding of the previous word $x_{i-1}$ as input. Formally, the word distribution $p(w_i)$ is computed as
\begin{align}
f^d_i &= \sigma(W^d_r [h_{i-1}, x_{i-1}, f'_i] + b^d_r) \\
o^d_i &= \sigma(W^d_o [h_{i-1}, x_{i-1}, f'_i] + b^d_i) \\
i^d_i &= \sigma(W^d_i [h_{i-1}, x_{i-1}, f'_i] + b^d_o) \\
\widetilde{h}_i &= \tanh(W^d_h [h_{i-1}, x_{i-1}, f'_i] + b^d_h) \\
c^d_i &= f^d_i c^d_{i-1} + i^d_i \widetilde{h}_i \\
h_i &= o^d_i \tanh(c^d_i) \\
p(w_i) &= softmax(W^2_p \tanh(W^1_p h_i + b^1_p) + b^2_p) \\
x_i &= A w_i,
\end{align}
where $f^d_i, o^d_i, i^d_i, \widetilde{h}_i, c^d_i, h_i, \in \R^{H}$ are the forget/output/input gates and cell input/hidden/output states. These gates and states are controlled by the past cell output state $h_{i-1}$, overview thought vector $f'_i$ as well as input symbol $x_i$. $W^d_r, W^d_o, W^d_i, W^d_h \in \R^{H \times (H + d)}$, $b^d_r, b^d_o, b^d_i, b^d_h \in \R^{H}$ are the decoder LSTM parameters. $A$ is the word embedding matrix, it transforms the one-hot vector $w_i$ into an embedding presentation $x_i$. $W^1_p \in \R^{E \times H}, W^2_p \in \R^{V \times E}, b^1_p \in \R^{E}, b^w_p \in \R^{V}$ are multiple-layer perceptron parameters, which is used to estimate a word distribution. We set $H=1000, E=300, V=9600$ in our experiments, note that $E$ is the embedding size and $V$ is the vocabulary size.
The number of parameters is around $H \times (H + F + E) \times 4 + E \times V + H \times E = 9.5M$.

\subsection{Details concerning the generation of visual concepts from pure text sentences}
For a given caption we manually generate 10 visual concepts (out of the 1000 set of visual concepts in our dictionary). We achieve this by firstly going through the sentence to extract all the words belonging to the dictionary. In the case of the "truth generator", we sample 10 words if we have more than 10 extracted concepts
or we append zeros if we have less. In the case of the "noisy generator", we firstly sample two noisy words randomly and then follow the previous procedure to get the additional 8 visual concepts. During data processing, we filtered out the sentences containing less than 4 concepts to make sure the number of "truth words" is at least twice as many as the added noise.

\subsection{Details Concerning Pre-trainable Parameters}
Our unsupervised learning approach can be applied to the parameters of the decoder/reviewer except the transformation matrices $W_h, W_m$ which take the fc7 features as input, and whose parameter size is $4096 \times 1000 \times 2 = 8.2M$. Since the total parameter size is around $2.8M + 9.5M + 8.2M = 20.5M$, that is to say that more than $60\%$ of the model can be pre-trained.  



\subsection{Implementation}
Our implementation uses Theano~\cite{bergstra2011theano} and Caffe~\cite{jia2014caffe} and is based on the code of~\cite{fang2015captions}~\footnote{\url{https://github.com/kelvinxu/arctic-captions}} and ~\cite{xu2015show}~\footnote{\url{https://github.com/s-gupta/visual-concepts}}. The code will be made available on github after publication of our work. Our models were trained on a Tesla K20Xm graphics card with 6G Bytes of memory.


\newpage
\section{Appendix - Visualization of Concept Attention \& Captions}
\begin{figure}[!h]
\begin{minipage}[b]{\dimexpr\textwidth-2\fboxsep-2\fboxrule\relax}
\centering
   \includegraphics[page=7,width=\width\linewidth]{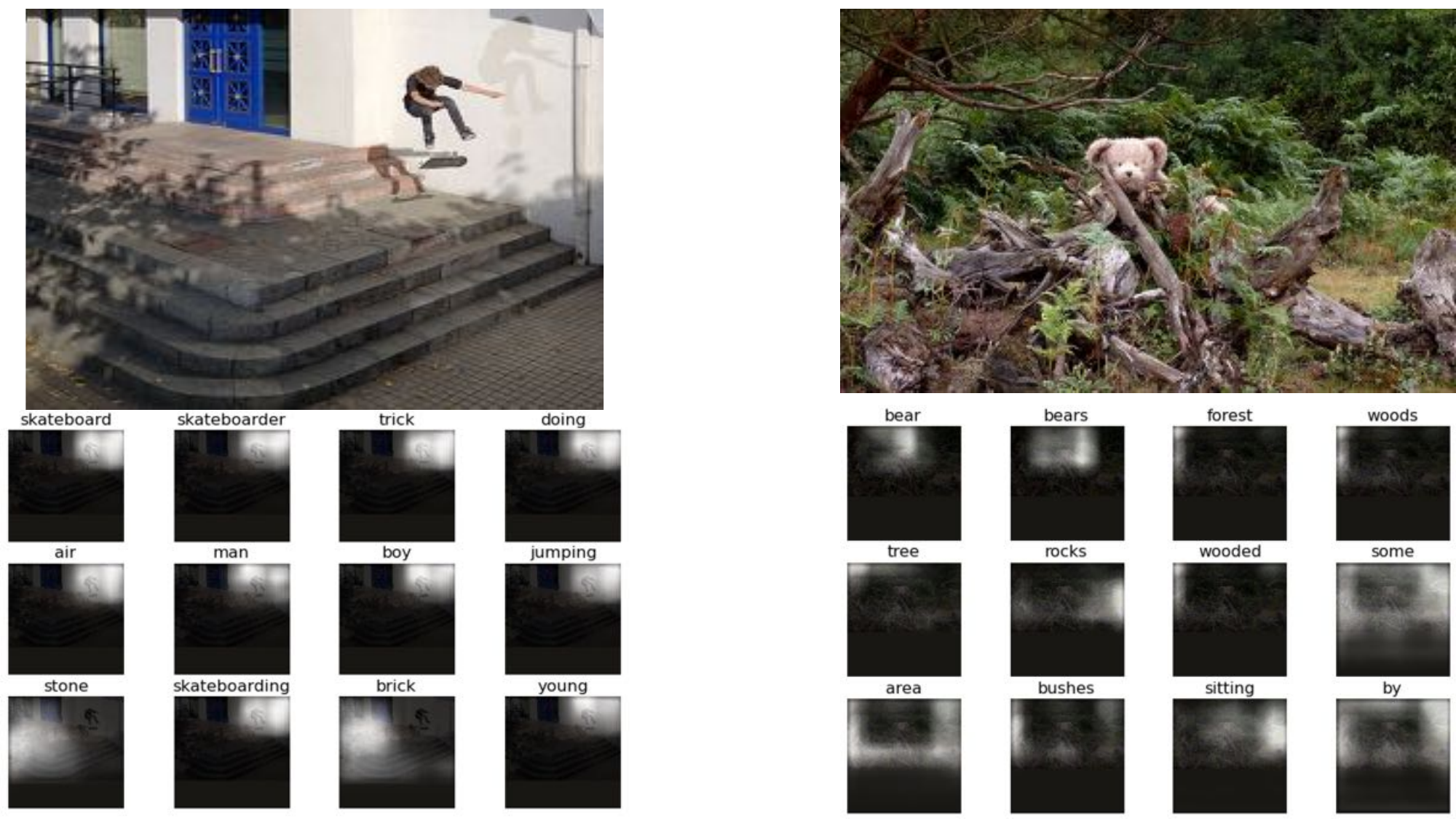}
\[ 
   \includegraphics[page=8,width=\width\linewidth]{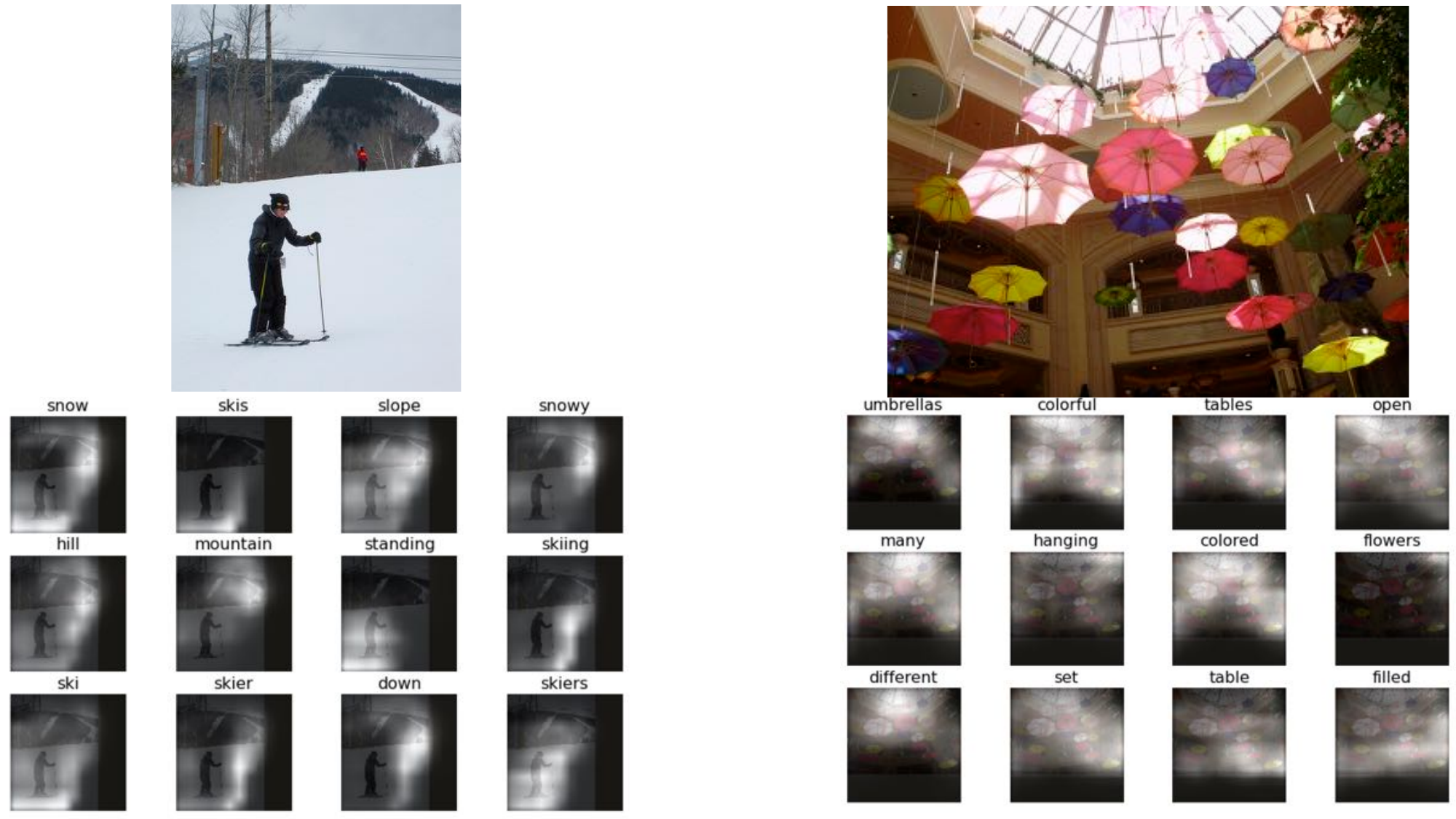}
\]
\end{minipage}
\caption{Additional examples of concept attention}
\end{figure}

\begin{figure}[ht]
\begin{minipage}{\dimexpr\textwidth-2\fboxsep-2\fboxrule\relax}
\centering
   \includegraphics[page=9,width=\width\linewidth]{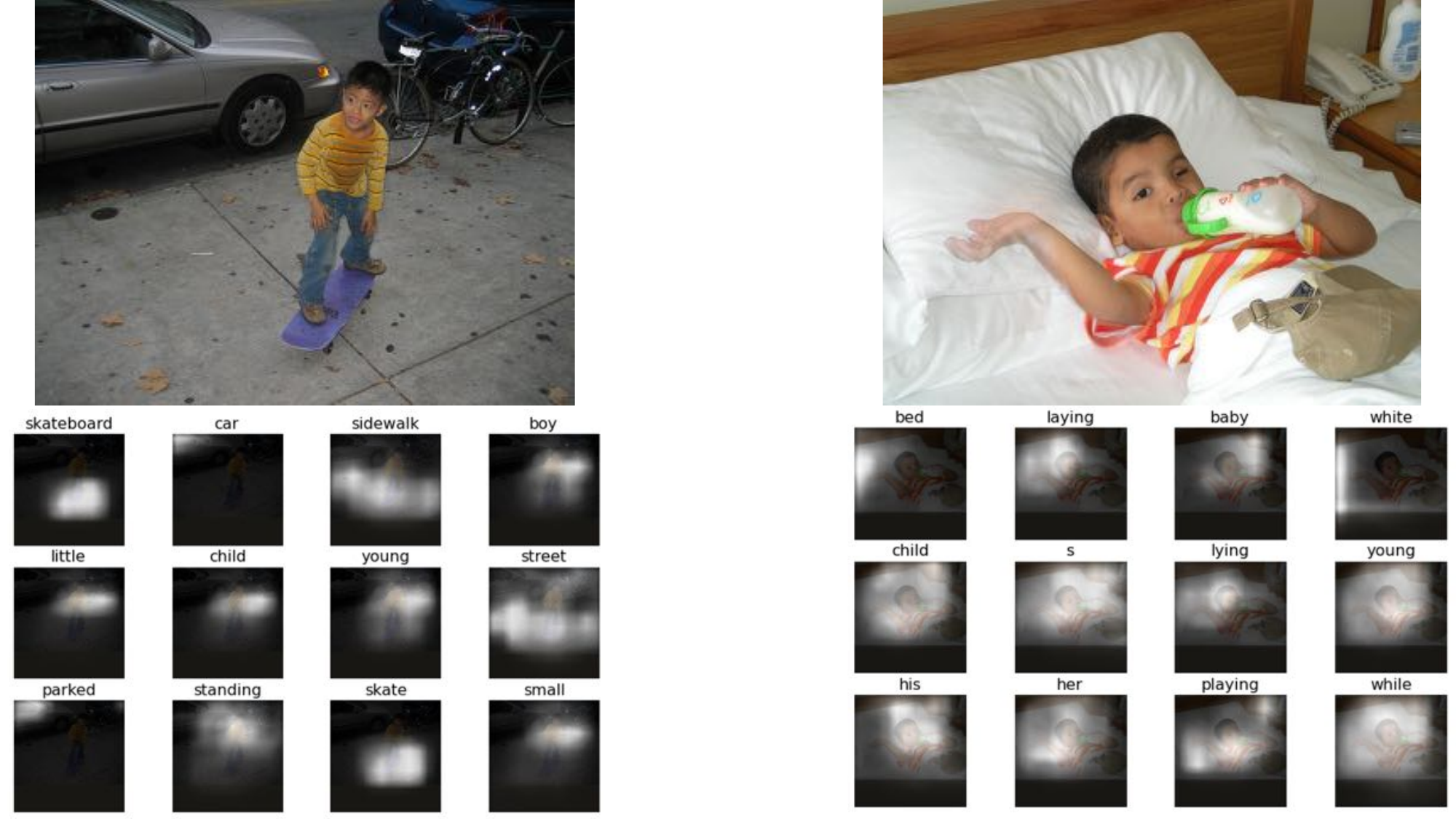}
\[    
   \includegraphics[page=10,width=\width\linewidth]{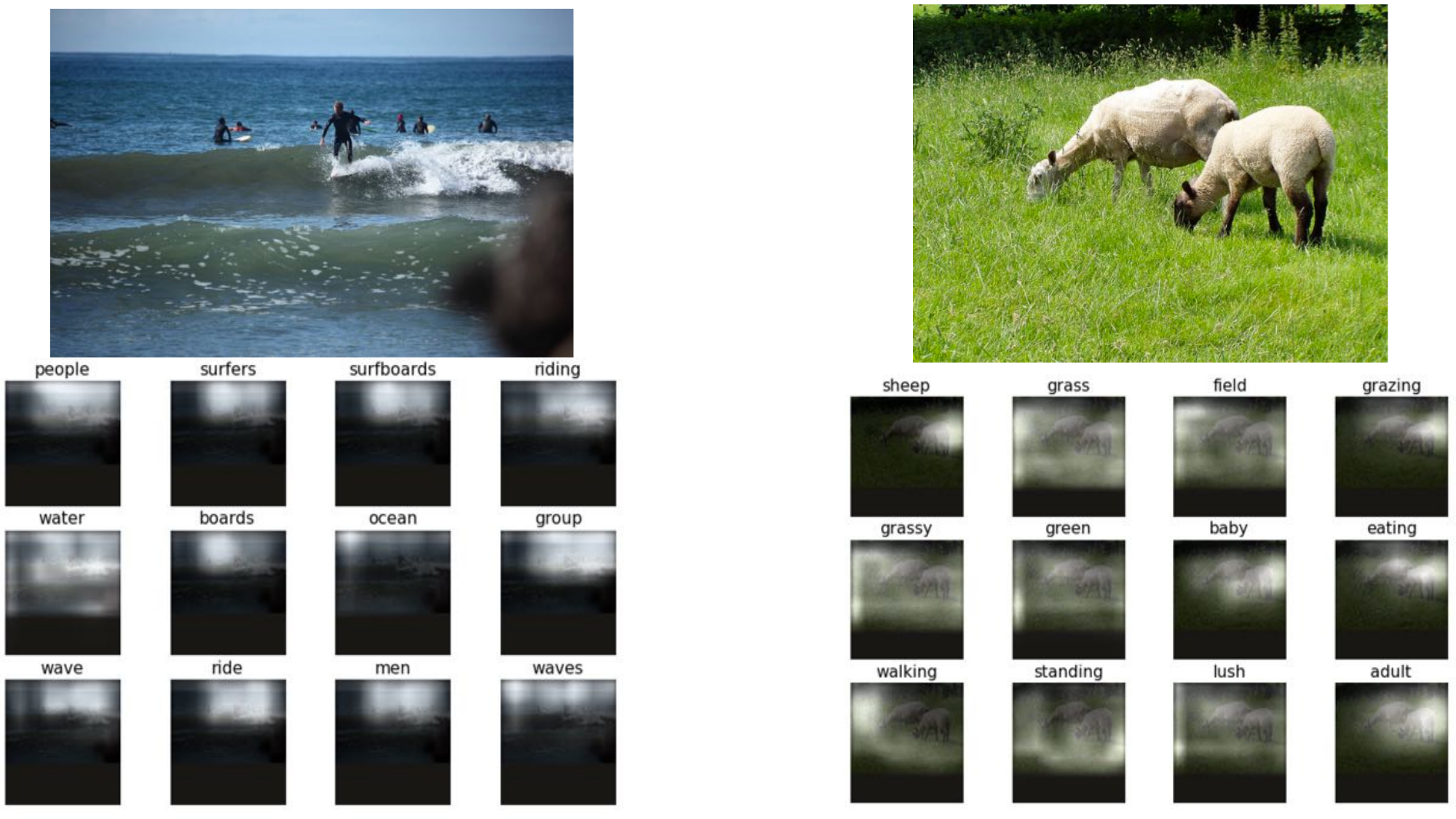}
\]
\end{minipage}
\caption{Additional examples of concept attention}
\end{figure}

\begin{figure}[htb]
\begin{minipage}{\dimexpr\textwidth-2\fboxsep-2\fboxrule\relax}
\centering
\includegraphics[page=1,width=\width\linewidth]{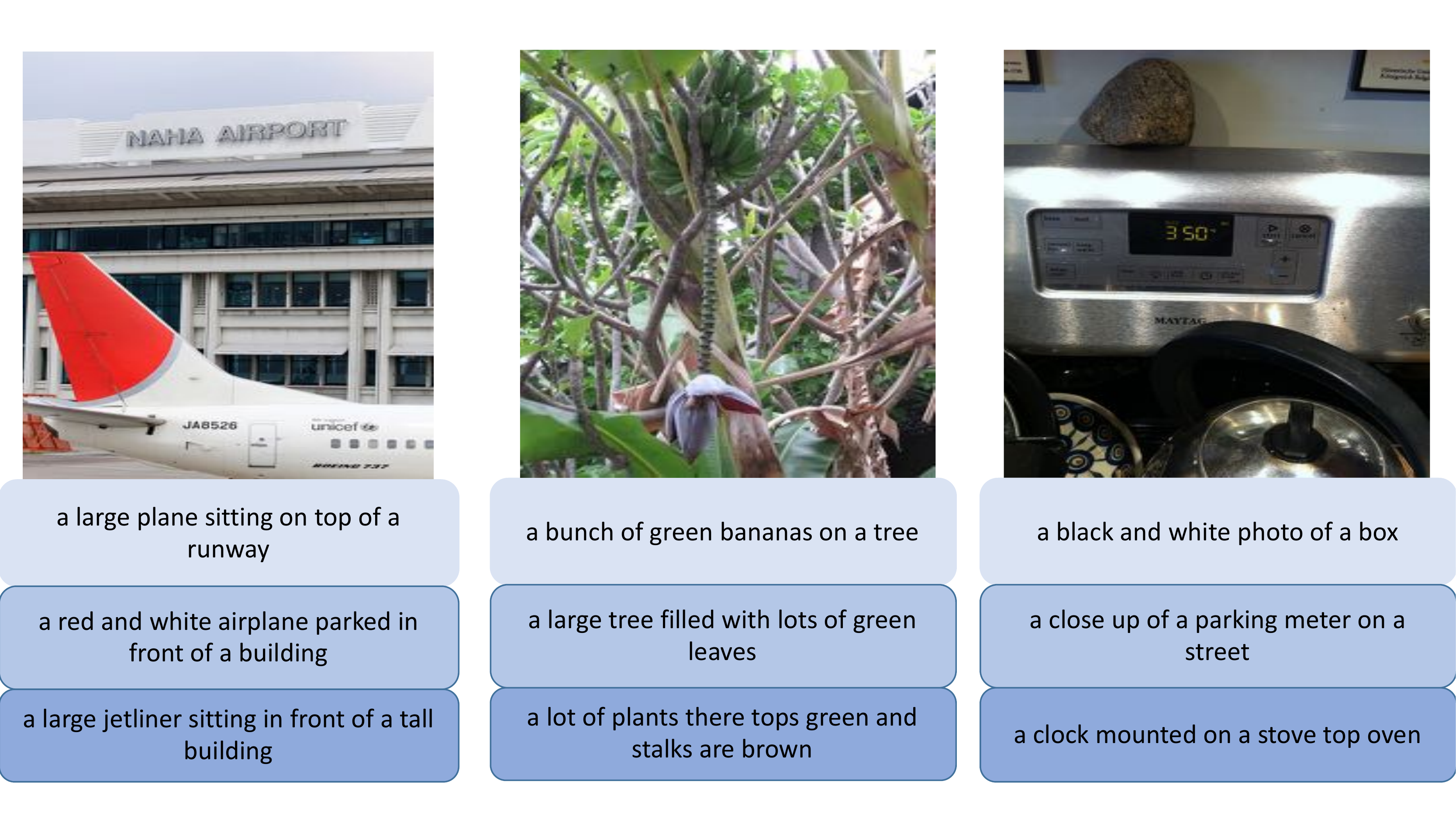}
\[
\includegraphics[page=2,width=\width\linewidth]{appendix}
\]
\end{minipage}
\caption{Additional examples of captions on the MS-COCO dataset. yellow: without pre-training, green: with pre-training, orange: groundtruth.}
\end{figure}

\begin{figure}[htb]
\begin{minipage}{\dimexpr\textwidth-2\fboxsep-2\fboxrule\relax}
\centering
\includegraphics[page=3,width=\width\linewidth]{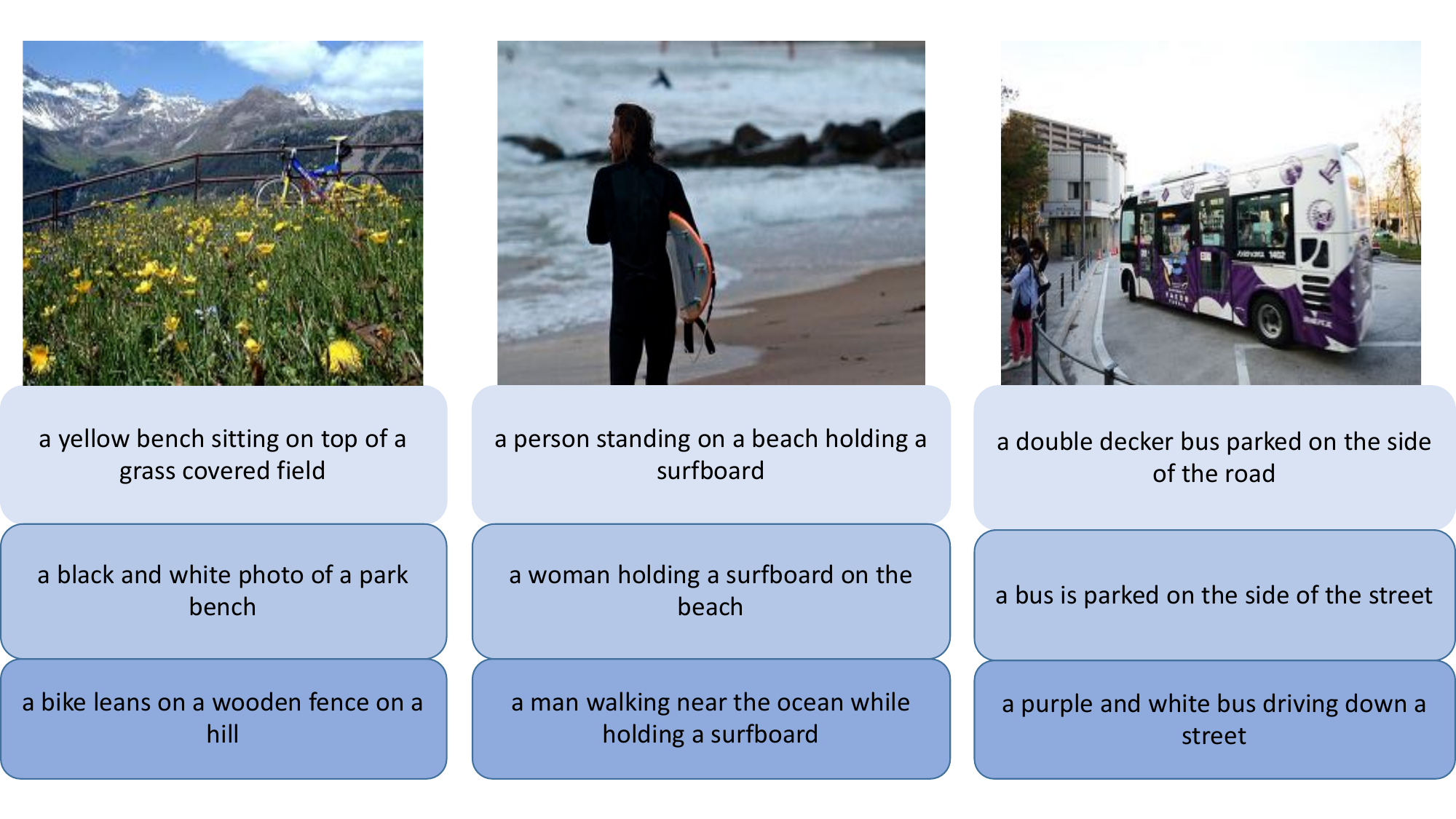}
\[
\includegraphics[page=4,width=\width\linewidth]{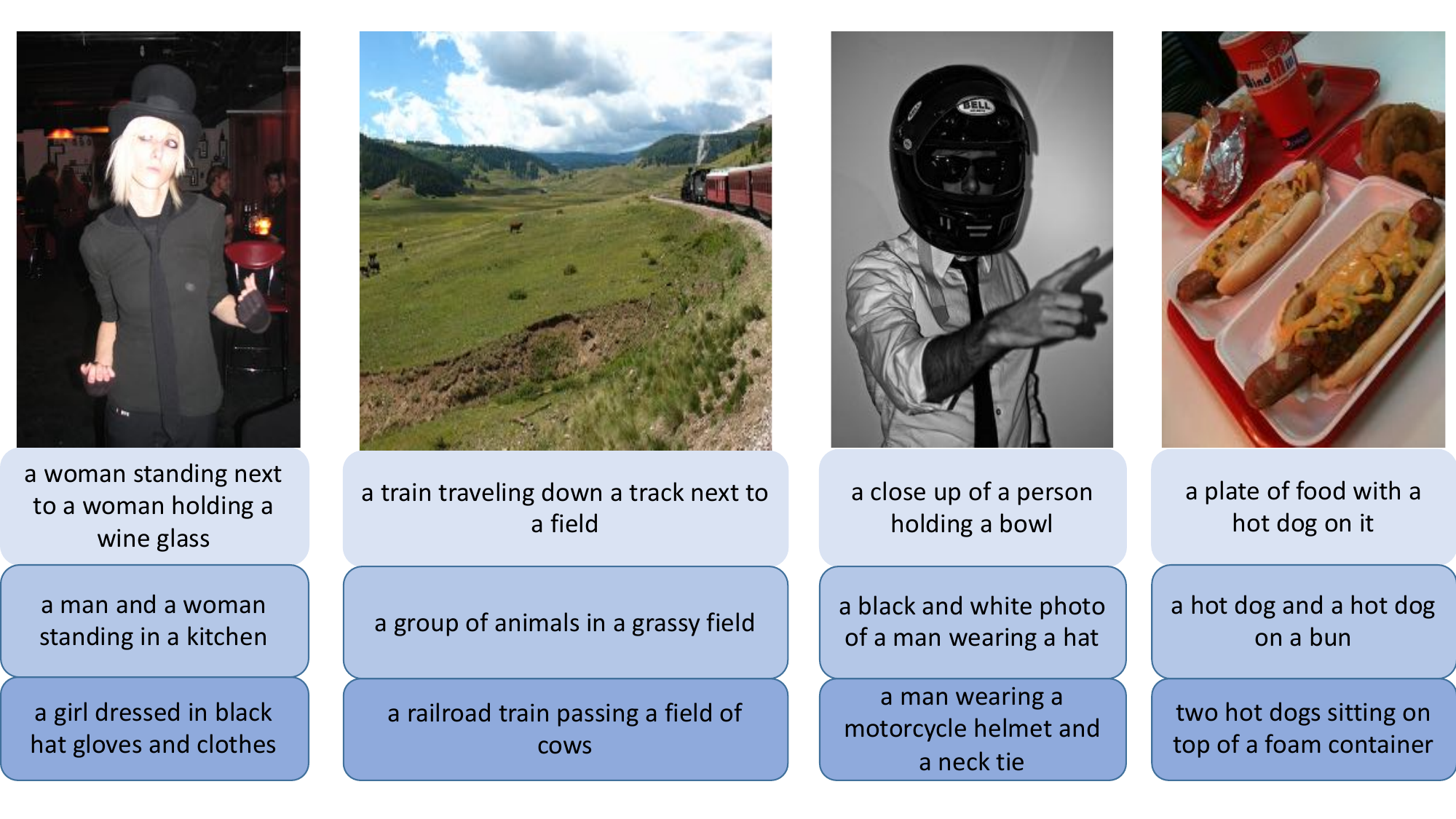}
\]
\end{minipage}
\caption{Additional examples of captions on the MS-COCO dataset. yellow: without pre-training, green: with pre-training, orange: groundtruth.}
\end{figure}

\begin{figure}[htb]
\begin{minipage}{\dimexpr\textwidth-2\fboxsep-2\fboxrule\relax}
\centering
\includegraphics[page=5,width=\width\linewidth]{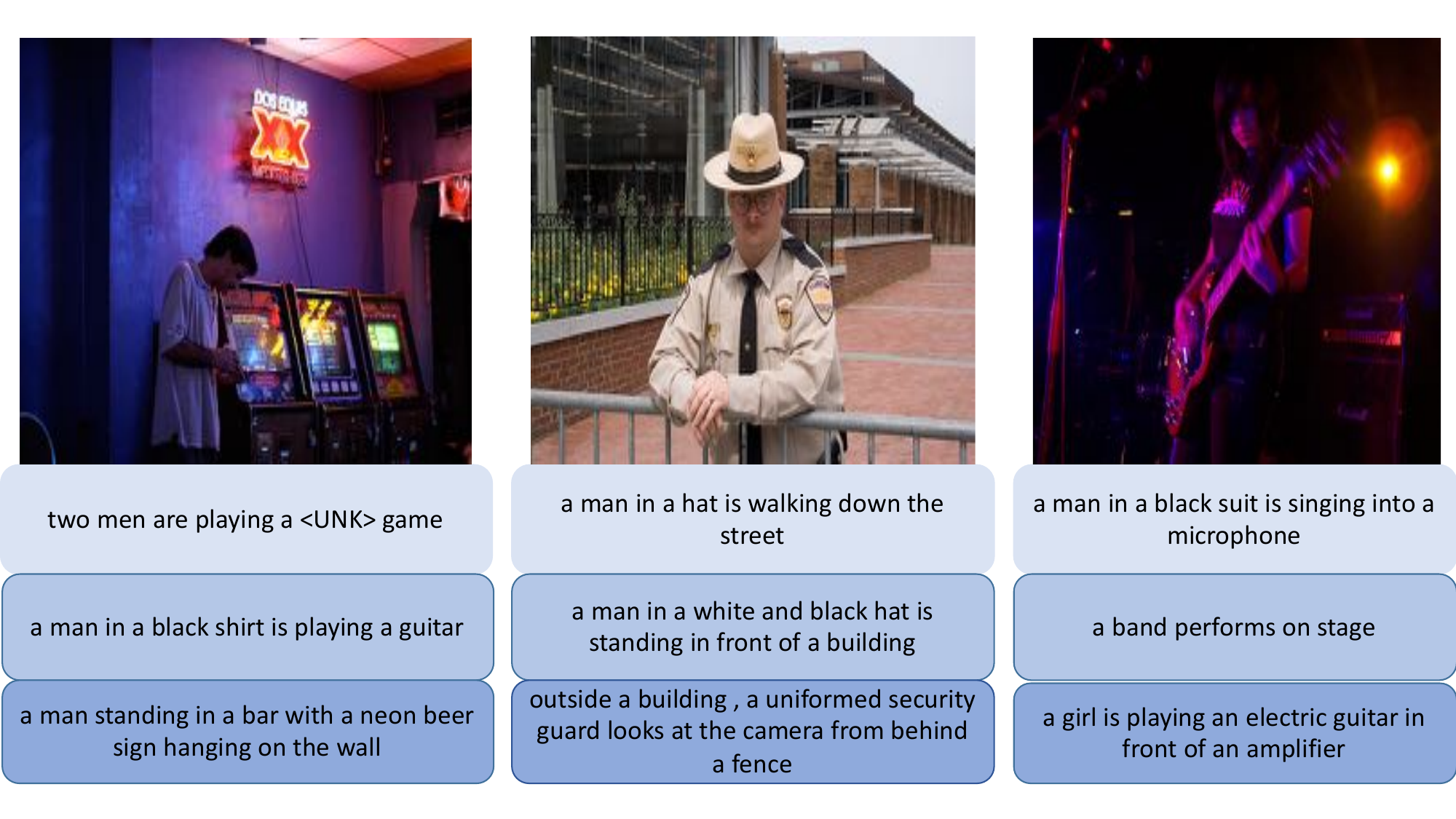}
\[
\includegraphics[page=6,width=\width\linewidth]{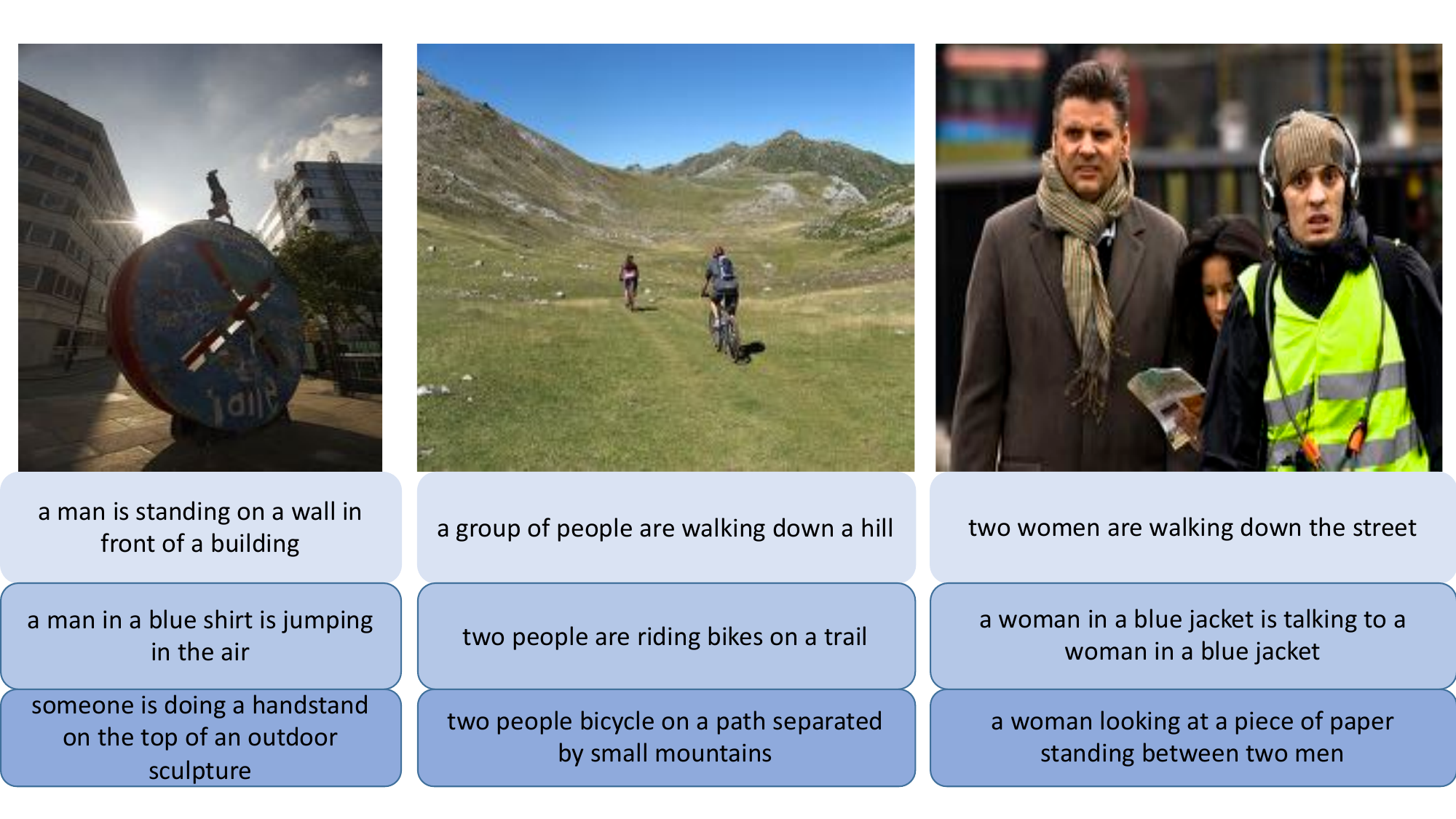}
\]
\end{minipage}
\caption{Additional examples of captions on the Flickr30K dataset. yellow: without pre-training, green: with pre-training, orange: groundtruth.}
\end{figure}